\documentclass{tlp}

\usepackage[T1]{fontenc}
\usepackage{amssymb}
\usepackage{amsmath}
\usepackage{epsfig}
\usepackage{theorem}
\usepackage{url}

\newtheorem{theorem}{Theorem}[section] 
\newtheorem{corollary}[theorem]{Corollary} 
\newtheorem{definition}[theorem]{Definition} 
\newtheorem{proposition}[theorem]{Proposition} 
\newtheorem{lemma}[theorem]{Lemma} 
\newtheorem{example}[theorem]{Example} 

\DeclareMathSymbol{\naf}{\mathord}{symbols}{"18}

\newcommand{\union}{\cup}
\newcommand{\isect}{\cap}
\newcommand{\Isect}{\bigcap}
\newcommand{\rg}[3]{{#1}{#2}\ldots{#2}{#3}}
\newcommand{\set}[1]{\{#1\}}
\newcommand{\sel}[2]{\{{#1}\mid{#2}\}}
\newcommand{\eset}[2]{\{{#1},\ldots,{#2}\}}
\newcommand{\pset}[1]{\mathbf{2}^{#1}}
\newcommand{\func}[2]{:{#1}\rightarrow{#2}}
\newcommand{\pair}[2]{\langle{#1},{#2}\rangle}
\newcommand{\tuple}[1]{\langle{#1}\rangle}

\newcommand{\compl}[1]{\overline{#1}}
\newcommand{\order}[1]{\mathcal{O}(#1)}

\newcommand{\IF}{\leftarrow}
\newcommand{\END}{.\ }
\newcommand{\at}[1]{\mathsf{#1}}
\newcommand{\lequiv}{\leftrightarrow}
\newcommand{\limpl}{\rightarrow}

\newcommand{\ren}[1]{#1^{\bullet}}
\newcommand{\renh}[1]{#1^{\circ}}
\newcommand{\cnt}[2]{#1\ \{#2\}}
\newcommand{\choice}[1]{\{#1\}}
\newcommand{\limit}[2]{{#1}\leq\{{#2}\}}
\newcommand{\compute}[1]{\mathsf{compute}\ \{#1\}}
\newcommand{\wsum}[2]{\mathrm{WS}_{#1}(#2)}

\newcommand{\hb}[1]{\mathrm{Hb}(#1)}
\newcommand{\hba}[1]{\mathrm{Hb_a}(#1)}
\newcommand{\hbv}[1]{\mathrm{Hb_v}(#1)}
\newcommand{\hbh}[1]{\mathrm{Hb_h}(#1)}
\newcommand{\hid}[1]{#1_{\mathrm{h}}}
\newcommand{\vis}[1]{#1_{\mathrm{v}}}

\newcommand{\trop}[1]{\mathrm{Tr}_{\mathrm{#1}}}
\newcommand{\tr}[2]{\mathrm{Tr}_{\mathrm{#1}}(#2)}
\newcommand{\len}[1]{||#1||}

\newcommand{\lm}[1]{\mathrm{LM}(#1)}
\newcommand{\sm}[1]{\mathrm{SM}(#1)}
\newcommand{\lfp}[1]{\mathrm{lfp}(#1)}
\newcommand{\nec}[1]{\mathrm{T}_{#1}}

\newcommand{\compst}[1]{\mathrm{CompS}(#1)}
\newcommand{\GLred}[2]{#1^{#2}}
\newcommand{\eval}[2]{#1/#2}
\newcommand{\simplify}[3]{\mathrm{simp}(#1,#2,#3)}

\newcommand{\eqt}{\mathrm{EQT}}
\newcommand{\trhid}[1]{\mathrm{Hidden}^{\circ}(#1)}
\newcommand{\trlm}[1]{\mathrm{Least}^{\bullet}(#1)}
\newcommand{\unstable}[1]{\mathrm{UnStable}(#1)}
\newcommand{\lpeq}[1]{\equiv_{\mathrm{#1}}}
\newcommand{\forget}[2]{\mathrm{fg}(#1,#2)}

\newcommand{\system}[1]{\textsc{#1}}
\newcommand{\prob}[1]{\textsf{#1}}

\newcommand{\kw}[1]{\text{#1}}
\newcommand{\proc}[1]{\mathsf{#1}}

\newcommand{\NP}{\mathbf{NP}}
\newcommand{\FNP}{\mathbf{FNP}}
\newcommand{\coNP}{\mathbf{co}\NP}
\newcommand{\coPH}[1]{\mathbf{\Pi^{\mathrm{p}}_{#1}}}

\newcommand{\lang}[1]{\mathbf{#1}}
\newcommand{\SAT}[1]{\mathrm{sat}(#1)}
\newcommand{\UNSAT}[1]{\mathrm{unsat}(#1)}
\newcommand{\sus}[2]{\mathrm{SU}_{#1}(#2)}
\newcommand{\ext}{\mathrm{Ext}}

\makeatletter
\def\@underjournal{%
  \vbox to 5.6\p@{\noindent\parbox[t]{4.8in}{\normalfont\affilsize\rmfamily
    {\itshape To appear in
     Theory and Practice of Logic Programming\/}\\[2.5\p@]
     \ }%
  \vss}%
}
\let\@j@urnal\@underjournal
\makeatother

\title[Theory and Practice of Logic Programming]
{Automated
Verification of Weak Equivalence \\ within the \system{smodels}
System\thanks{This is an extended version of a paper
\protect\cite{JO02:jelia} presented at the 8th European Workshop on
Logics in Artificial Intelligence in Cosenza, Italy.}}

\author[T. Janhunen and E. Oikarinen]
{TOMI JANHUNEN and EMILIA OIKARINEN \\
 Helsinki University of Technology \\
 Department of Computer Science and Engineering \\
 Laboratory for Theoretical Computer Science \\
 P.O. Box 5400, FI-02015 TKK, Finland \\
 \email{Tomi.Janhunen@tkk.fi, Emilia.Oikarinen@tkk.fi}}

\submitted{7 December 2005}
\revised{30 June 2006}
\accepted{23 August 2006}

\begin{document}

\maketitle


\begin{abstract}
In answer set programming (ASP), a problem at hand is solved by (i)
writing a logic program whose answer sets correspond to the solutions
of the problem, and by (ii) computing the answer sets of the program
using an \emph{answer set solver} as a search engine.  Typically, a
programmer creates a series of gradually improving logic programs for
a particular problem when optimizing program length and execution time
on a particular solver. This leads the programmer to a meta-level
problem of ensuring that the programs are equivalent, i.e., they give
rise to the same answer sets.
To ease answer set programming at methodological level, we propose a
translation-based method for verifying the equivalence of logic
programs. The basic idea is to translate logic programs $P$ and $Q$
under consideration into a single logic program $\eqt(P,Q)$ whose
answer sets (if such exist) yield counter-examples to the equivalence
of $P$ and $Q$. The method is developed here in a slightly more
general setting by taking the {\em visibility} of atoms properly into
account when comparing answer sets. The translation-based approach
presented in the paper has been implemented as a translator called
\system{lpeq} that enables the verification of weak equivalence within
the \system{smodels} system using the same search engine as for the
search of models. Our experiments with \system{lpeq} and
\system{smodels} suggest that establishing the equivalence of logic
programs in this way is in certain cases much faster than naive
cross-checking of answer sets.
\end{abstract}



\section{Introduction}


Answer set programming (ASP) has recently been proposed and promoted
as a self-standing logic programming paradigm
\cite{MT99:slp,Niemela99:amai,GL02:aij}.
Indeed, the paradigm has received increasing attention since efficient
implementations such as \system{dlv} \cite{LPFEGPS06:acmtocl} and
\system{smodels} \cite{SNS02:aij} became available in the late
nineties. There are numerous applications of ASP ranging, e.g., from
product configuration \cite{SNTS01:asp} to a decision support system
of the space shuttle \cite{BBGBW01:padl}. The variety of answer set
solvers is also rapidly growing as new solvers are being developed
constantly for the sake of efficiency. The reader
is referred to
\cite{JNSY00:kr,LZ02:aaai,LM04:lpnmr,Janhunen04:ecai,AGLNS05:lpnmr,%
GJMSTT05:lpnmr,LT05:lpnmr}
in this respect.


Despite the declarative nature of ASP, the development of programs
resembles that of programs in conventional programming. That is, a
programmer often develops a series of gradually improving programs for
a particular problem, e.g., when optimizing execution time and space.
As a consequence, the programmer needs to ensure that subsequent
programs which differ in performance yield the same output.
This setting leads us to the problem of verifying whether given two
logic programs $P$ and $Q$ have exactly the same answer sets, i.e.,
are {\em weakly equivalent} (denoted $P\lpeq{}Q$). Looking at this
from the ASP perspective, weakly equivalent programs produce the same
solutions for the problem that they formalize.


There are also other notions of equivalence that have been proposed
for logic programs. \citeN{LPV01:acmtocl} consider $P$ and $Q$ {\em
strongly equivalent}, denoted $P\lpeq{s}Q$, if and only if $P\union
R\lpeq{}Q\union R$ for all programs $R$ each of which acts as a
potential context for $P$ and $Q$. By setting $R=\emptyset$ in the
definition of $\lpeq{s}$, we obtain that $P\lpeq{s}Q$ implies
$P\lpeq{}Q$ but the converse does not hold in general.
Consequently, the question whether $P\lpeq{}Q$ holds remains open
whenever $P\not\lpeq{s}Q$ turns out to be the case. This implies that
verifying $P\lpeq{}Q$ remains as a problem of its own, which cannot be
fully compensated by verifying $P\lpeq{s}Q$.
As suggested by its name, $\lpeq{s}$ is a much stronger relation than
$\lpeq{}$ in the sense that the former relates far fewer programs than
the latter. This makes $\lpeq{s}$ better applicable to {\em
subprograms} or {\em program modules} constituting larger programs
rather than complete programs for which $\lpeq{}$ is more natural.
Moreover, there is a number of characterizations of strong equivalence
\cite{LPV01:acmtocl,PTW01:epia,Lin02:kr,Turner03:tplp} which among
other things indicate that strongly equivalent programs are
classically equivalent, but not necessarily vice versa as to be
demonstrated in Example
\ref{ex:classically-but-not-strongly-equivalent}.
Thus strong equivalence permits only classical {\em program
transformations}, i.e., substitutions of a program module (a set of
rules) by another. In contrast to this, weak equivalence is more
liberal as regards program transformations some of which are not
classical but still used in practice; the reader may consult Example
\ref{ex:weakly-but-not-strongly} for an instance.


For the reasons discussed above, we concentrate on the case of
complete programs and weak equivalence in this article.  We develop a
method that extends \cite{JO02:jelia,JO04:lpnmr} and hence fully
covers the class of {\em weight constraint programs} supported by the
front-end \system{lparse} \cite{LPARSE01:manual} used with the
\system{smodels} system \cite{SNS02:aij}.
The key idea in our approach is to {\em translate} logic programs $P$
and $Q$ under consideration into a single logic program $\eqt(P,Q)$
which has an answer set if and only if $P$ has an answer set that is
not an answer set of $Q$. Such answer sets, if found, act as {\em
counter-examples} to the equivalence of $P$ and $Q$. Consequently, the
equivalence of $P$ and $Q$ can be established by showing that
$\eqt(P,Q)$ and $\eqt(Q,P)$ have no answer sets.%
\footnote{Turner \cite{Turner03:tplp} develops an analogous
transformation for weight constraint programs and {\em strong
equivalence}. Moreover, \citeN{EFTW04:lpnmr} cover the case of
disjunctive programs under strong and {\em uniform equivalence} and
present the respective transformations.}
Thus the existing search engine of the \system{smodels} system can be
used for the search of counter-examples and there is no need to
develop a special purpose search engine for the verification task.
Moreover, we are obliged to develop the underlying theory in a more
general setting where programs may involve {\em invisible} atoms,
e.g., generated by \system{lparse} when compiling weight constraints.
The basic idea is that such atoms should be neglected by equivalence
relations but this is not the case for $\lpeq{}$ and $\lpeq{s}$.
To this end, we apply yet another equivalence relation, namely {\em
visible equivalence} denoted by $\lpeq{v}$
\cite{Janhunen03:report,Janhunen06:jancl}.
This relation is compatible with $\lpeq{}$ in the sense that these
equivalence relations coincide in the absence of invisible atoms. In
fact, we develop a translation-based verification method for
$\lpeq{v}$ and characterize the class of \system{smodels} programs for
which the method is guaranteed to work by constraining the use of
invisible atoms. This class is identified as the class of programs
possessing {\em enough visible atoms}. Most importantly, this property
is shared by weight constraint programs produced by the front-end
\system{lparse} during grounding.

The rest of this paper is organized as follows.
The rule-based syntax of logic programs supported by the current
\system{smodels} system is described in Section \ref{section:syntax}.
It is then explained in Section \ref{section:semantics} how the
semantics of such rules is covered by the stable model semantics
proposed by Gelfond and Lifschitz \citeyear{GL88:iclp}.
Section \ref{section:equivalence} introduces the notion of {\em
visible equivalence} mentioned above. We perform a preliminary
complexity analysis of the problem of verifying $P\lpeq{v}Q$ for $P$
and $Q$ given as input. Unfortunately, recent complexity results
\cite{ETW05:ijcai} suggest discouraging rises of complexity in the
presence of invisible atoms. Thus we need to impose additional
constraints in order to keep the verification problem in
$\coNP$; thus enabling the use of \system{smodels} as search
engine in a feasible way.
In Section~\ref{section:translation}, we present our translation-based
method for verifying the visible equivalence of \system{smodels}
programs. The correctness of the method is also addressed.
The resulting complexity classifications are
then concluded in Section \ref{section:complexity}.
Section \ref{section:weightconstraint} concentrates on the case of
weight constraint programs supported by the front-end \system{lparse} of
the \system{smodels} system and shows how programs in the extended
language are covered by the translation-based method.
Section \ref{section:experiments} is devoted to experiments that we
have performed with an implementation of the translation-based method,
a translator called \system{lpeq}, and the \system{smodels} system.
The results indicate that in certain cases verifying the equivalence
of \system{smodels} programs using \system{lpeq} is one or two orders
of magnitude faster than naive cross-checking of stable models.
Finally, the paper is finished by a brief conclusion in Section
\ref{section:conclusion}.


\section{Programs in the \system{smodels} Language}
\label{section:syntax}

The goal of this section is to make the reader acquainted with the
rule-based language supported by the current \system{smodels} system
\cite{SNS02:aij}. Definition \ref{def:rules} lists five forms of rules
which constitute the knowledge representation primitives of the
system.
Besides {\em basic rules} (\ref{eq:basic-rule}) of conventional {\em
normal logic programs}, there are also other expressions such as
{\em constraint rules} (\ref{eq:constraint-rule}), {\em choice rules}
(\ref{eq:choice-rule}), {\em weight rules} (\ref{eq:weight-rule}),
and {\em compute statements} (\ref{eq:compute-statement}).
These extensions have been carefully chosen to be directly and
efficiently implementable in the search engine of the \system{smodels}
system \cite{SNS02:aij}. It should be stressed that the front-end of
the system, \system{lparse} \cite{LPARSE01:manual}, admits a more
liberal use of constraint and weight rules
\cite{Syrjanen04:jelia}
but we postpone the discussion of such features until
Section~\ref{section:weightconstraint}.

\begin{definition}
\label{def:rules}
Rules are expressions of the forms
\begin{eqnarray}
\label{eq:basic-rule}
h\IF\rg{a_1}{,}{a_n},\rg{\naf b_1}{,}{\naf b_m} \\
\label{eq:constraint-rule}
h\IF\cnt{c}{\rg{a_1}{,}{a_n},\rg{\naf b_1}{,}{\naf b_m}} \\
\label{eq:choice-rule}
\choice{\rg{h_1}{,}{h_l}}\IF\rg{a_1}{,}{a_n},\rg{\naf b_1}{,}{\naf b_m} \\
\label{eq:weight-rule}
h\IF\limit{w}{\rg{a_1=w_{a_1}}{,}{a_n=w_{a_n}},
              \rg{\naf b_1=w_{b_1}}{,}{\naf b_m=w_{b_m}}} \\
\label{eq:compute-statement}
\compute{\rg{a_1}{,}{a_n},\rg{\naf b_1}{,}{\naf b_m}}
\end{eqnarray}
where $n\geq 0$, $m\geq 0$, and $l>0$, and where $h$, each $a_i$,
each $b_j$, and each $h_k$ are atoms and $c$, each $w_{a_i}$, each
$w_{b_j}$, as well as $w$, are natural numbers.
\end{definition}

The symbol $\naf$ occurring in Definition \ref{def:rules} denotes {\em
default negation} or {\em negation as failure to prove} which differs
from classical negation in an important way \cite{GL90:iclp}.
We define positive and negative {\em default literals}
in the standard way as atoms $a$ or their negations $\naf a$,
respectively.
The exact model-theoretic semantics of rules is deferred until Section
\ref{section:semantics}, but --- informally speaking --- the rules
listed above are used to draw conclusions as follows.

\begin{itemize}
\item
The head $h$ of a basic rule (\ref{eq:basic-rule}) can be inferred if
the atoms $\rg{a_1}{,}{a_n}$ are inferable by other rules whereas the
atoms $\rg{b_1}{,}{b_m}$ are {\em not}.

\item
The head $h$ of a constraint rule (\ref{eq:constraint-rule}) can be
inferred if the number of inferable atoms among $\rg{a_1}{,}{a_n}$
plus the number of non-inferable atoms among $\rg{b_1}{,}{b_m}$ is at
least $c$.

\item
A choice rule (\ref{eq:choice-rule}) is similar to a basic rule except
that any subset of the non-empty set of head atoms $\eset{h_1}{h_l}$
can be inferred instead of a single head atom $h$. Note that it is not
necessary to infer any of the head atoms.

\item
A weight rule (\ref{eq:weight-rule}) involves summing as follows: the
weight $w_{a_i}$ (resp.\ $w_{b_j}$) is one of the summands if and only
if $a_i$ is inferable (resp.\ $b_j$ is not inferable). The head $h$
can be inferred if such a sum of weights is at least $w$.

\item
The default literals involved in a compute statement
(\ref{eq:compute-statement}) act as direct constraints saying that the
atoms $\rg{a_1}{,}{a_n}$ should be inferable by some rules whereas the
atoms $\rg{b_1}{,}{b_m}$ should not.
\end{itemize}

A couple of observations follows.
A constraint rule (\ref{eq:constraint-rule}) becomes equivalent to a
basic rule (\ref{eq:basic-rule}) given that $c=n+m$. A weight rule
(\ref{eq:weight-rule}) reduces to a constraint rule
(\ref{eq:constraint-rule}) when all weights are equal to $1$ and
$w=c$. Moreover, default literals may be assigned different weights in
different weight rules, i.e., weights are local in this sense.
The types of rules defined above are already well-suited for a variety
of knowledge representation and reasoning tasks in a number of
domains. Example \ref{ex:cafe} demonstrates the use of rules in a
practical setting. The reader is referred to
\cite{Niemela99:amai,MT99:slp,SNS02:aij,GL02:aij}
for more examples how to represent knowledge in terms of rules.

\begin{example}
\label{ex:cafe}
Consider the task of describing coffee orders using
rules\footnote{Rules are separated with full stops and the symbol
``$\IF$'' is dropped from a basic rule (\ref{eq:basic-rule}) or a
choice rule (\ref{eq:choice-rule}) if the {\em body} of the rule is
empty ($n=0$ and $m=0$).} introduced in Definition \ref{def:rules}.
The nine rules given below form our formalization of this domain
which should be self-explanatory. The compute statement in the
end identifies the orders of interest to be those for which
``$\at{acceptable}$'' can be inferred.
\begin{center}
\begin{tabular}{l}
$\choice{\at{coffee},\at{tea},\at{biscuit},\at{cake},\at{cognac}}$. \\
$\choice{\at{cream},\at{sugar}}\IF\at{coffee}$. \\
$\at{cognac}\IF\at{coffee}$. \\
$\choice{\at{milk},\at{lemon},\at{sugar}}\IF\at{tea}$. \\
$\at{mess}\IF\at{milk},\at{lemon}$. \\
$\at{happy}\IF\cnt{1}{\at{biscuit},\at{cake},\at{cognac}}$. \\
$\at{bankrupt}\IF
 \limit{6}{\at{coffee}=1,\at{tea}=1,
           \at{biscuit}=1,\at{cake}=2,\at{cognac}=4}$. \\
$\at{acceptable}\IF\at{happy},\naf\at{bankrupt},\naf\at{mess}$. \\
$\compute{\at{acceptable}}$.
\end{tabular}
\end{center}
\end{example}

We define a {\em logic program} $P$ as a finite\footnote{This reflects
the fact that the theory being presented/developed here is closely
related to an actual implementation, the \system{smodels} engine,
which admits only finite sets of ground rules.} set of ground rules of
the forms (\ref{eq:basic-rule})--(\ref{eq:compute-statement}) given in
Definition \ref{def:rules}. It follows that programs under
consideration are fully instantiated and thus consist of ground atoms
which are parallel to \emph{propositional atoms}, or \emph{atoms} for
short in the sequel.
The {\em Herbrand base} of a logic program $P$ can be any fixed set of
atoms $\hb{P}$ containing all atoms that actually appear in the rules
of $P$. Furthermore, we view $\hb{P}$ as a part of the program
which corresponds to defining a logic program as pair
$\pair{P}{\hb{P}}$ where $\hb{P}$ acts as the symbol table of $P$.
The flexibility of this definition has important consequences.  First,
the \emph{length} $\len{P}$ of the program, i.e., the number of
symbols needed to represent $P$ as a string, becomes dependent on
$|\hb{P}|$.  This aspect becomes relevant in the analysis of
translation functions \cite{Janhunen06:jancl}. Second, the explicit
representation of $\hb{P}$ enables one to keep track of atoms whose
occurrences have been removed from a program, e.g., due to program
optimization. For instance, the program $\pair{\set{a\IF\naf
b\END}}{\set{a,b}}$ can be rewritten as
$\pair{\set{a\END}}{\set{a,b}}$ under stable model semantics.

There is a further aspect of atoms that affects the way we
treat Herbrand bases, namely the {\em visibility} of atoms. It is
typical in answer set programming that only certain atoms appearing in
a program are relevant for representing the solutions of the problem
being solved. Others act as auxiliary concepts that might not appear
in other programs written for the same problem. As a side effect, the
models/interpretations assigned to two programs may differ already on
the basis of auxiliary atoms. Rather than introducing an explicit hiding
mechanism in the language itself, we let the programmer decide the visible
part of $\hb{P}$, i.e., $\hbv{P}\subseteq\hb{P}$ which determines the
set of {\em hidden} atoms $\hbh{P}=\hb{P}-\hbv{P}$. The
ideas presented so far are combined as follows.

\begin{definition}
\label{def:logic-program}
A logic program in the \system{smodels} system (or an \system{smodels}
program for short) is a triple $\tuple{P,\hbv{P},\hbh{P}}$ where
\begin{enumerate}
\item
$P$ is a finite set of rules of the forms
(\ref{eq:basic-rule}) -- (\ref{eq:compute-statement});

\item
$\hbv{P}$ and $\hbh{P}$ are finite and disjoint sets of atoms and
determine the visible and hidden Herbrand bases of the program,
respectively; and

\item
all atoms occurring in $P$ are contained in $\hb{P}=\hbv{P}\union\hbh{P}$.
\end{enumerate}
Finally, we define $\hba{P}$ as the set of atoms of $\hb{P}$ not
occurring in $P$.\footnote{The atoms in $\hba{P}$ are made false by
stable semantics to be introduced in Section \ref{section:semantics}.}
\end{definition}

Note that the atoms of $\hba{P}$ can be viewed as {\em additional}
atoms that just extend $\hb{P}$.
By a slight abuse of notation, we often use $P$ rather than the whole
triple when referring to a program $\tuple{P,\hbv{P},\hbh{P}}$.  To
ease the treatment of programs, we make some default assumptions
regarding the sets $\hb{P}$ and $\hbv{P}$.  Unless otherwise stated,
we assume that $\hbv{P}=\hb{P}$, $\hbh{P}=\emptyset$, and
$\hba{P}=\emptyset$, i.e., $\hb{P}$ contains only atoms that actually
appear in $P$.

\begin{example}
Given $P=\set{a\IF\naf b\END}$, the default
interpretation is that $\hb{P}=\set{a,b}$,
$\hbv{P}=\hb{P}=\set{a,b}$, and $\hbh{P}=\emptyset$.
To make an exception in this respect, we have to add explicitly that,
e.g., $\hbv{P}=\set{a,c}$ and $\hbh{P}=\set{b}$.
Together with $P$ these declarations imply that
$\hba{P}$ is implicitly assigned to $\set{c}$.
\end{example}

Generally speaking, the set $\hbv{P}$ can be understood as a {\em program
interface} of $P$ and it gives the basis for comparing the program $P$
with other programs of interest. The atoms in $\hbh{P}$ are to be
hidden in any such comparisons.


\section{Stable Model Semantics}
\label{section:semantics}

In this section, we review the details of {\em stable model semantics}
proposed by \citeN{GL88:iclp}. Stable models were first introduced in
the context of {\em normal logic programs}, i.e., logic programs that
solely consist of basic rules (\ref{eq:basic-rule}), but soon they
were generalized for other classes involving syntactic extensions.
In addition to recalling the case of normal programs, it is also
important for us to understand how the semantic principles underlying
stable models can be applied to the full syntax of \system{smodels}
programs introduced in Section \ref{section:syntax}.  Yet another
generalization will be presented in Section
\ref{section:weightconstraint} where the class of weight constraint
programs is addressed.

The class of normal programs includes {\em positive programs} that are
free of default negation, i.e., $m=0$ for all rules
(\ref{eq:basic-rule}) of such programs.  The standard way to determine
the semantics of any positive program $P$ is to take the {\em least
model} of $P$, denoted by $\lm{P}$, as the semantical basis
\cite{Lloyd87}. This is a particular classical model of $P$ which is
minimal with respect to subset\footnote{It is assumed that
interpretations are represented as sets of atoms evaluating to true.}
inclusion and also unique with this property. Moreover, the least
model $\lm{P}$ coincides with the intersection of all classical models
of $P$. Consequently, an atom $a\in\hb{P}$ is a logical consequence of
$P$ in the classical sense if and only if $a\in\lm{P}$. It is also
important to realize that the semantic operator $\lm{\cdot}$ is
inherently monotonic: $P\subseteq Q$ implies $\lm{P}\subseteq\lm{Q}$
for any positive normal logic programs $P$ and $Q$.

\citeN{GL88:iclp} show how the least model semantics can be
generalized to cover normal logic programs.  The idea is to reduce a
normal logic program $P$ with respect to a model candidate $M$ by
pre-interpreting negative literals that appear in the rules of $P$.
The resulting program $\GLred{P}{M}$ --- also known as the
Gelfond-Lifschitz {\em reduct} of $P$ --- contains a reduced rule
$h\IF\rg{a_1}{,}{a_n}$ if and only if there is a rule
(\ref{eq:basic-rule}) in $P$ so that the negative literals $\rg{\naf
b_1}{,}{\naf b_m}$ in the body are satisfied in $M$. This makes
$\GLred{P}{M}$ a positive program whose semantics is determined in the
standard way, i.e., using its least model \cite{Lloyd87}.

\begin{definition}[\citeN{GL88:iclp}]
\label{def:normal-sm}
For a normal logic program $P$, an interpretation $M\subseteq\hb{P}$
is a stable model of $P$ if and only if $M=\lm{\GLred{P}{M}}$.
\end{definition}

For a positive program $P$, the reduct $\GLred{P}{M}=P$ for any
$M\subseteq\hb{P}$ implying that $\lm{P}$ coincides with the unique
stable model of $P$. Unlike this, stable models need not be unique in
general: a normal logic program $P$ may possess several stable models
or no stable model at all. However, this is not considered as a
problem in answer set programming, since the aim is to capture
solutions to the problem at hand with the stable models of a program
that is constructed to formalize the problem. In particular, if there
are no solutions for the problem, then the logic programming
representation is not supposed to possess any stable models.

\citeN{Simons99:lpnmr} shows how the stable model semantics can be
generalized for the other kinds of rules presented in Section
\ref{section:syntax}. However, the reduced program is not explicitly
present in the semantical definitions given by him. This is why we
resort to an alternative definition, which appears as Definition
\ref{def:reduct} below. It will be explained in Section
\ref{section:weightconstraint} how the forthcoming definition can be
understood as a special case of that given by \citeN{SNS02:aij} for
more general classes of rules.  In contrast to their definitions that
involve {\em deductive closures} of sets of rules, we define stable
models purely in model-theoretic terms using the least model concept.

Given a logic program $P$, an {\em interpretation} $I$ is simply a
subset of $\hb{P}$ defining which atoms $a$ are considered to be {\em
true} ($a\in I$) and which {\em false} ($a\not\in I$). By the
following definition, we extend the satisfaction relation $I\models r$
for the types of rules $r$ under consideration. In particular, let us
point out that negative default literals are treated classically at
this point.

\begin{definition}
\label{def:satisfaction}
Given an interpretation $I\subseteq\hb{P}$ for an \system{smodels}
program $P$,

\begin{enumerate}
\item
A positive default literal $a$ is satisfied in $I$
(denoted $I\models a$) $\iff$ $a\in I$.

\item
A negative default literal $\naf a$ is satisfied in $I$
(denoted $I\models\naf a$) $\iff$ $I\not\models a$.

\item
A set of default literals $L$ is satisfied in $I$ (denoted $I\models L$)
$\iff$ \\ $I\models l$ for every $l\in L$.

\item
A basic rule $r$ of the form (\ref{eq:basic-rule}) is satisfied in $I$
(denoted $I\models r$)
$\iff$ \\
$I\models\set{\rg{a_1}{,}{a_n},\rg{\naf b_1}{,}{\naf b_m}}$
implies $I\models h$.

\item
A constraint rule $r$ of the form (\ref{eq:constraint-rule})
is satisfied in $I$ (denoted $I\models r$)
$\iff$ \\
$c\leq |\sel{a_i}{I\models a_i}\union\sel{\naf b_j}{I\models\naf b_j}|$
implies $I\models h$.

\item
A choice rule $r$ of the form (\ref{eq:choice-rule}) is always
satisfied in $I$.

\item
A weight rule $r$ of the form (\ref{eq:weight-rule})
is satisfied in $I$ (denoted $I\models r$)
$\iff$
\begin{equation}
\renewcommand{\arraystretch}{1.5}
\begin{array}{rcl}
w & \leq &
\wsum{I}{\rg{a_1=w_{a_1}}{,}{a_n=w_{a_n}},
         \rg{\naf b_1=w_{b_1}}{,}{\naf b_m=w_{b_m}}} \\
& = &
\displaystyle
\sum_{I\models a_i}w_{a_i}+\sum_{I\models\naf b_j}w_{b_j}
\end{array}
\label{eq:sum-of-weights}
\end{equation}
implies $I\models h$.

\item
A compute statement $s$ of the form (\ref{eq:compute-statement})
is satisfied in $I$ (denoted $I\models s$)
$\iff$
$I\models\set{\rg{a_1}{,}{a_n},\rg{\naf b_1}{,}{\naf b_m}}$.

\item
A program $P$ is satisfied in $I$ ($I\models P$)
$\iff$ $I\models r$ for every $r\in P$.
\end{enumerate}
\end{definition}

The equality in (\ref{eq:sum-of-weights}) determines how weighted
literal sets are evaluated.  Given an interpretation $I$ and an
assignment of weights to default literals as in the body of a weight
rule (\ref{eq:weight-rule}), the respective \emph{weight sum} in
(\ref{eq:sum-of-weights}) includes the weight of each literal true in
$I$.  This primitive will be needed a lot in the sequel to deal with
weight rules.

\begin{example}
The third but last rule of Example \ref{ex:cafe} is satisfied
in an interpretation
$I_1=\set{\at{tea},\at{biscuit}}$,
but not in
$I_2=\set{\at{coffee},\at{cake},\at{cognac}}$.
\end{example}

An interpretation $I$ is a (classical) {\em model} of a logic
program $P$ if and only if $I\models P$.  However, stable models are
not arbitrary models of logic programs.  As discussed in the beginning
of this section, they involve a reduction of logic programs which is
based on a pre-interpretation of negative literals.

\begin{definition}
\label{def:reduct}
For an \system{smodels} program $P$ and an interpretation
$I\subseteq\hb{P}$ of $P$, the reduct $\GLred{P}{I}$ contains
\begin{enumerate}
\item
a basic rule $h\IF\rg{a_1}{,}{a_n}$
$\iff$
there is a basic rule (\ref{eq:basic-rule}) in $P$ such that
$I\models\eset{\naf b_1}{\naf b_m}$
\textbf{or}
there is a choice rule (\ref{eq:choice-rule}) in $P$ such that
$h\in\eset{h_1}{h_l}$, $I\models h$, and $I\models\eset{\naf b_1}{\naf b_m}$;

\item
a constraint rule $h\IF\cnt{c'}{\rg{a_1}{,}{a_n}}$
$\iff$
there is a constraint rule (\ref{eq:constraint-rule}) in $P$ and
$c'=\max(0,c-|\sel{\naf b_i}{I\models \naf b_i}|)$;

\item
a weight rule $h\IF\limit{w'}{\rg{a_1=w_{a_1}}{,}{a_n=w_{a_n}}}$
$\iff$
there is a weight rule (\ref{eq:weight-rule}) in $P$ and
$w'=
 \max(0,w-\wsum{I}{\rg{\naf b_1=w_{b_1}}{,}{\naf b_m=w_{b_m}}})$; and

\item
no compute statements.
\end{enumerate}
\end{definition}

Note that in addition to evaluating negative literals in the bodies of
rules, the head atoms $h\in\eset{h_1}{h_l}$ of choice rules
(\ref{eq:choice-rule}) are subject to a special treatment: an
essential prerequisite for including $h\IF\rg{a_1}{,}{a_n}$ in the
reduct $\GLred{P}{M}$ is that $M\models h$, i.e., $h\in M$. This is
the way in which the choice regarding $h$ takes place.
Moreover, it is clear by Definition \ref{def:reduct} that the reduct
$\GLred{P}{M}$ is free of default negation and it contains only basic
rules, constraint rules, and weight rules, but no compute statements.
Thus we call an \system{smodels} program $P$ {\em positive} if each rule
$r\in P$ is of the forms (\ref{eq:basic-rule}), (\ref{eq:constraint-rule})
and (\ref{eq:weight-rule}) restricted to the case $m=0$.
The least model semantics can be generalized for positive programs by
distinguishing their minimal models.

\begin{definition}
\label{def:minimal-model}
A model $M\models P$ of a (positive) \system{smodels} program $P$ is
{\em minimal} if and only if there is no $M'\models P$
such that $M'\subset M$.
\end{definition}

Positive programs share many important properties of positive {\em
normal programs} and the straightforward semantics based on minimal
models and the least model is easily generalized for positive
programs.

\begin{definition}
\label{def:true-operator}
For a positive \system{smodels} program $P$, we define an operator
$\nec{P}\func{\pset{\hb{P}}}{\pset{\hb{P}}}$
as follows. Given any interpretation $I\subseteq\hb{P}$, the result of
applying $\nec{P}$ to $I$, i.e., $\nec{P}(I)\subseteq\hb{P}$, contains
an atom $a\in\hb{P}$ if and only if
\begin{enumerate}
\item
there is a basic rule $a\IF\rg{a_1}{,}{a_n}\in P$ and
$I\models\eset{a_1}{a_n}$; or

\item
there is a constraint rule $a\IF\cnt{c}{\rg{a_1}{,}{a_n}}\in P$ and
$c\leq|\sel{a_i}{I\models a_i}|$; or

\item
there is a weight rule $a\IF\limit{w}{\rg{a_1=w_{a_1}}{,}{a_n=w_{a_n}}}\in P$
and
\[w\leq\wsum{I}{\rg{a_1=w_{a_1}}{,}{a_n=w_{a_n}}}.\]
\end{enumerate}
\end{definition}

Intuitively, the operator $\nec{P}$ gives atoms that are necessarily
true by the rules of $P$ if the atoms in $I$ are assumed to be true.
It follows that $\nec{P}(I)\subseteq I$ implies $I\models P$ in
general. We are now ready to state a number of properties of
positive programs.

\begin{proposition}
\label{prop:least-model}
Let $P$ be a positive \system{smodels} program.
\begin{enumerate}
\item
For any collection $C$ of models of $P$, the intersection $\Isect C$
is also a model of $P$.

\item
The program $P$ has a unique minimal model $M$, i.e., the
{\em least model} $\lm{P}$ of $P$.

\item
The least model $\lm{P}=\Isect\sel{I\subseteq\hb{P}}{I\models P}$
and $\lm{P}=\lfp{\nec{P}}$.
\end{enumerate}
\end{proposition}

Moreover, positive programs are {\em monotonic} in the sense
that $P_1\subseteq P_2$ implies $\lm{P_1}\subseteq\lm{P_2}$.
Given the least model semantics for positive programs, it becomes
straightforward to generalize the stable model semantics
\cite{GL88:iclp} for programs involving default negation. The key idea
is to use the reduction from Definition \ref{def:reduct}, but the
effect of compute statements must also be taken into account as they
are dropped out by Definition \ref{def:reduct}. To this end, we define
$\compst{P}$ as the union of literals appearing in the compute
statements (\ref{eq:compute-statement}) of $P$.

\begin{definition}
\label{def:sm}
An interpretation $M\subseteq\hb{P}$ is a stable model of an
\system{smodels} program $P$ if and only if $M=\lm{\GLred{P}{M}}$ and
$M\models\compst{P}$.
\end{definition}

Definition \ref{def:sm} reveals the purpose of compute statements:
they are used to select particular models among those satisfying the
conventional fixed point condition from Definition \ref{def:normal-sm}.
Given any logic program $P$, we define the set
\begin{equation}
\sm{P}=
\sel{M\subseteq\hb{P}}{M=\lm{\GLred{P}{M}}\text{ and }M\models\compst{P}}.
\end{equation}
In analogy to the case of normal logic programs, the number of stable
models may vary in general. A positive program $P$ has a unique stable
model $\lm{P}$ as $\GLred{P}{M}=P$ holds; recall that compute
statements are not allowed in positive programs.  It is also worth
noting that $M=\lm{\GLred{P}{M}}$ and $M\models\compst{P}$ imply
$M\models P$, i.e., stable models are also classical models in the
sense of Definition \ref{def:satisfaction}.
However, the converse does not hold in general, i.e., $M\models P$
need not imply $M=\lm{\GLred{P}{M}}$ although it certainly implies
$M\models\compst{P}$. For example, interpretations $M_1=\set{a}$ and
$M_2=\set{a,b}$ are models of the program $P=\set{a\IF\cnt{1}{\naf
a,\naf b}\END}$, but only $M_1$ is stable. To verify this, note that
$\GLred{P}{M_1}=\set{a\IF\cnt{0}{}\END}$ and
$\GLred{P}{M_2}=\set{a\IF\cnt{1}{}\END}$.

\begin{example}
\label{ex:cafe2}
Recall the program $P$ from Example \ref{ex:cafe}. According to
\system{smodels} there are 33 acceptable orders that are captured by
the stable models of $P$. One of them is
$M_7=
 \{\at{acceptable},\at{happy},
   \at{lemon},\at{tea},\at{biscuit}\}$.
The reader is kindly asked to verify $M_7=\lm{\GLred{P}{M_7}}$ and
$M_7\models\compst{P}$ using the reduct $\GLred{P}{M_7}$ listed below.
\begin{center}
\begin{tabular}{l}
$\at{tea}$. $\at{biscuit}$. \\
$\at{cognac}\IF\at{coffee}$. \\
$\at{lemon}\IF\at{tea}$. \\
$\at{mess}\IF\at{milk},\at{lemon}$. \\
$\at{happy}\IF\cnt{1}{\at{biscuit},\at{cake},\at{cognac}}$. \\
$\at{bankrupt}\IF
 \limit{6}{\at{coffee}=1,\at{tea}=1,\at{biscuit}=1,\at{cake}=2,\at{cognac}=4}$.
\\
$\at{acceptable}\IF\at{happy}$.
\end{tabular}
\end{center}
\end{example}


\section{Notions of Equivalence}
\label{section:equivalence}

We begin this section by reviewing two fundamental notions of
equivalence that have been proposed for logic programs, namely {\em
weak} and {\em strong} equivalence, and point out some of their
limitations.
This is why we resort to another notion of equivalence in Section
\ref{section:visible-equivalence}: {\em visible} equivalence is a
variant of weak equivalence which takes the visibility of atoms better
into account. Then we are ready to identify the respective
verification problem in Section \ref{section:verification-problem} and
discuss in which way invisible atoms render the verification problem
more difficult. This serves as a starting point for characterizing a
subclass of programs for which visible equivalence can be verified
using a translation-based technique in analogy to \cite{JO02:jelia}.

\citeN{LPV01:acmtocl} address two major notions of equivalence for
logic programs. The first one arises naturally from the stable model
semantics.

\begin{definition}
Logic programs $P$ and $Q$ are {\em weakly equivalent}, denoted
$P\lpeq{}Q$, if and only if $\sm{P}=\sm{Q}$, i.e., $P$ and $Q$ have
the same stable models.
\end{definition}

The second notion is definable in terms of the first and the
definition is given relative to a class of logic programs which is
represented by $R$ below. Of course, a natural choice for us would be
the class of \system{smodels} programs but that is not made explicit
in the following definition.

\begin{definition}
Logic programs $P$ and $Q$ are {\em strongly equivalent}, denoted
$P\lpeq{s}Q$, if and only if $P\union R\lpeq{}Q\union R$ for any logic
program $R$.
\end{definition}

Here the program $R$ can be understood as an arbitrary context in
which the other two programs $P$ and $Q$ being compared could be
placed. This is how strongly equivalent logic programs can be used as
semantics preserving substitutes of each other. This feature makes
$\lpeq{s}$ a {\em congruence relation} over the class of logic
programs under consideration: if $P\lpeq{s}Q$ holds, then also
$P\union R\lpeq{s}Q\union R$ holds for any $R$.
Moreover, it is easy to see that $P\lpeq{s}Q$ implies $P\lpeq{}Q$, but
not necessarily vice versa: $\lpeq{s}$ relates far fewer programs than
$\lpeq{}$ as demonstrated in Example \ref{ex:weakly-but-not-strongly}.
This explains why we call $\lpeq{}$ the {\em weak equivalence}
relation for the class of logic programs introduced in Sections
\ref{section:syntax} and \ref{section:semantics}. It is worth pointing
out that whereas $\lpeq{}$ is an equivalence relation it does not
permit substitutions ($P\lpeq{}Q$ does not imply $P\union
R\lpeq{}Q\union R$ in general) and hence it does not qualify as a
congruence relation.

\begin{example}
\label{ex:weakly-but-not-strongly}
Consider $P=\set{a\IF\naf b\END}$ and $Q=\set{a\END}$.
It is easy to see that $\sm{P}=\sm{Q}=\set{\set{a}}$ and
$P\lpeq{}Q$.
However, when joined with $R=\set{b\END}$, we note that $\sm{P\union
R}\neq\sm{Q\union R}$ holds so that $P\not\lpeq{s}Q$.
The programs $P$ and $Q$ are not classically equivalent either as
$M\models P$ and $M\not\models Q$ hold for $M=\set{b}$.
\end{example}

Although the relation $\lpeq{s}$ appears attractive at first glance, a
drawback is that it is quite restrictive, allowing only rather
straightforward semantics-preserving transformations of (sets of) rules.
In fact, \citeN{LPV01:acmtocl} characterize $\lpeq{s}$ in Heyting's
logic {\em here-and-there} (HT) which is an intermediary logic between
intuitionistic and classical propositional logics.
This result implies that each program transformation admitted by
$\lpeq{s}$ is based on a classical equivalence of the part being
replaced (say $P$) and its substitute (say $Q$), i.e., $P\lpeq{s}Q$
implies that $P$ and $Q$ are classically equivalent. However, the
converse is not true in general as there are classically equivalent
programs that are not strongly equivalent.

\begin{example}
\label{ex:classically-but-not-strongly-equivalent}
The propositional sentence $a\lequiv(\neg a\limpl a)$ is classically
valid --- suggesting a program transformation that replaces
$P=\set{a\END}$ by $Q=\set{a\IF\naf a\END}$. However, since
$\sm{P}=\set{\set{a}}$ and $\sm{Q}=\emptyset$, we have $P\not\lpeq{}Q$
and $P\not\lpeq{s}Q$ although $P$ and $Q$ are classically equivalent.
\end{example}

Since $\lpeq{s}$ is a congruence relation, it is better applicable to
{\em subprograms} or {\em program modules} constituting larger
programs rather than complete programs. In contrast to this, weak
equivalence is mainly targeted to the comparison of complete programs
in terms of their stable models. Due to the nature of ASP, this is
often the ultimate question confronted by a programmer when optimizing
and debugging programs. For this reason, we concentrate on the problem
of verifying weak equivalence in this paper and we leave the
modularization aspects of weak and visible equivalence to be addressed
elsewhere
\cite{OJ06:ecai}.

\subsection{Visible Equivalence}
\label{section:visible-equivalence}

We do not find the notion of weak equivalence totally satisfactory
either. For $P\lpeq{}Q$ to hold, the stable models in $\sm{P}$ and
$\sm{Q}$ have to be identical subsets of $\hb{P}$ and $\hb{Q}$,
respectively. This makes $\lpeq{}$ less useful if $\hb{P}$ and
$\hb{Q}$ differ by some (in)visible atoms which are not trivially
false in all stable models. As already discussed in Section
\ref{section:syntax}, such atoms are needed when some auxiliary
concepts are formalized using rules. The use of such atoms/concepts
may lead to more concise encodings of problems as demonstrated by our
next example.

\begin{example}
\label{ex:even-subsets}
Consider the following programs consisting of basic rules, choice rules,
and compute statements. The parameter $n$ below is an odd natural number.

\begin{center}
\begin{tabular}{ll}
Program $P_n$:
&$\choice{\at{bit}_1,\at{bit}_2,\ldots,\at{bit}_n}.$ \\
&$\at{odd}\IF\at{bit}_1,\naf\at{bit}_2,\ldots,\naf\at{bit}_n.$ \\
&$\at{odd}\IF\naf\at{bit}_1,\at{bit}_2,\naf\at{bit}_3\ldots,\naf\at{bit}_n.$ \\
&$\vdots$ \\
&$\at{odd}\IF\at{bit}_1,\at{bit}_2,\at{bit}_3,
             \naf\at{bit}_4\ldots,\naf\at{bit}_n.$ \\
&$\at{odd}\IF\naf\at{bit_1},\at{bit}_2,\at{bit}_3,\at{bit}_4,
             \naf\at{bit}_5\ldots,\naf\at{bit}_n.$ \\
&$\vdots$ \\
&$\at{odd}\IF\at{bit}_1,\ldots,\at{bit}_n.$ \\
&$\compute{\naf\at{odd}}.$ \\
\ \\
Program $Q_n$:
&$\choice{\at{bit}_1,\at{bit}_2,\ldots,\at{bit}_n}.$ \\
&$\at{odd}_1\IF\at{bit}_1.$ \\
&$\at{odd}_2\IF\at{bit}_2,\naf\at{odd}_1.$
 $\at{odd}_2\IF\naf\at{bit}_2,\at{odd}_1.$ \\
&$\vdots$ \\
&$\at{odd}_n\IF\at{bit}_n,\naf\at{odd}_{n-1}.$
 $\at{odd}_n\IF\naf\at{bit}_n,\at{odd}_{n-1}.$ \\
&$\at{odd}\IF\at{odd}_n.$ \\
&$\compute{\naf\at{odd}}.$ \\
\end{tabular}
\end{center}

The first program generates all subsets $B$ of
$BIT_n=\choice{\at{bit}_1,\at{bit}_2,\ldots,\at{bit}_n}$, analyzes
when $|B|$ is odd, and accepts only subsets with non-odd (even)
cardinality. Thus $P_n$ has $2^{n-1}$ stable models $M\subseteq BIT_n$
with $|M|$ even but also $2^{n-1}$ basic rules capturing subsets with
odd cardinality.
In contrast, huge savings can be achieved by introducing new atoms
$\at{odd}_1,\ldots,\at{odd}_n$ so that each $\at{odd}_i$ is supposed
to be true if and only if
$|B\isect\set{\at{bit}_1,\ldots,\at{bit}_i}|$ is odd. Using these, the
oddness of $|B|$ can be formalized in terms of $2n$ basic rules. The
resulting program $Q_n$ has $2^{n-1}$ stable models, but they are not
identical with the stable models of $Q_n$ due to new atoms
involved. Thus we have $\sm{Q_n}\neq\sm{P_n}$ and
$Q_n\not\lpeq{}P_n$ for every odd natural number $n$.
\end{example}

From the programmer's point of view, the programs $P_n$ and $Q_n$
solve the same problem and should be considered equivalent if one
neglects the interpretations of $\at{odd}_1,\ldots,\at{odd}_n$ in the
stable models of $Q_n$. To this end, we adopt a slightly more general
notion of equivalence \cite{Janhunen03:report,Janhunen06:jancl} which
takes the visibility of atoms properly into account. The key idea is
that when two programs $P$ and $Q$ are compared, the hidden atoms in
$\hbh{P}$ and $\hbh{Q}$ are considered to be local to $P$ and $Q$ and
thus negligible as far as the equivalence of the programs is
concerned. In addition to this feature, a very strict (bijective)
correspondence of stable models is necessitated by the notion of
\emph{visible equivalence}.

\begin{definition}
\label{def:visible-equivalence}
Logic programs $P$ and $Q$ are {\em visibly equivalent}, denoted
$P\lpeq{v}Q$, if and only if $\hbv{P}=\hbv{Q}$ and there is a
bijection $f\func{\sm{P}}{\sm{Q}}$ such that for every $M\in\sm{P}$,
$M\isect\hbv{P}=f(M)\isect\hbv{Q}$.
\end{definition}

\begin{proposition}
The relation $\lpeq{v}$ is an equivalence relation.
\end{proposition}

By defining $\hbv{P_n}=\hbv{Q_n}=BIT_n$ for the programs $P_n$ and
$Q_n$ defined in Example \ref{ex:even-subsets} we obtain an intuitive
relationship $Q_n\lpeq{v}P_n$. The bijection $f$ involved in this
relationship maps a stable model $M\in\sm{Q_n}$ to another
$f(M)=M\isect BIT_n\in\sm{P_n}$.
Our following example demonstrates the case in which both $\sm{P}$ and
$\sm{Q}$ have stable models that cannot be distinguished if projected
to $\hbv{P}=\hbv{Q}$, i.e., there are stable models $M,N\in\sm{P}$
such that $M\isect\hbv{P}=N\isect\hbv{P}$ and analogously for $Q$.
However, this does not necessarily exclude the possibility for a
bijection in the sense of Definition \ref{def:visible-equivalence}.

\begin{example}
\label{ex:visible-equivalence}
Consider logic programs
$P=\set{a\IF b\END
        a\IF c\END
        b\IF\naf c\END
        c\IF\naf b\END}$
and
$Q=\set{\choice{b,c}\END 
        a\IF b,c\END
        a\IF\naf b,\naf c\END
        b\IF c,\naf b\END
        c\IF b,\naf c\END}$
with $\hbv{P}=\hbv{Q}=\set{a}$ and $\hbh{P}=\hbh{Q}=\set{b,c}$.  The
stable models of $P$ are $M_1=\set{a,b}$ and $M_2=\set{a,c}$ whereas
for $Q$ they are $N_1=\set{a}$ and $N_2=\set{a,b,c}$. Thus
$P\not\lpeq{}Q$ is clearly the case, but we have a bijection
$f\func{\sm{P}}{\sm{Q}}$, which maps $M_i$ to $N_i$ for
$i\in\set{1,2}$, such that $M\isect\hbv{P}=f(M)\isect\hbv{Q}$.
Thus $P\lpeq{v}Q$ holds.
\end{example}

A brief comparison of $\lpeq{v}$ and $\lpeq{}$ follows.

\begin{proposition}
\label{prop:weak-meets-visible}
If $\hb{P}=\hb{Q}$ and $\hbh{P}=\hbh{Q}=\emptyset$,
then $P\lpeq{}Q$ $\iff$ $P\lpeq{v}Q$.
\end{proposition}

In words, the two relations coincide when all atoms are visible.
There is only a slight difference: $\lpeq{v}$ insists on
$\hb{P}=\hb{Q}$ whereas $\lpeq{}$ does not.  Nevertheless, it follows
by Definition \ref{def:logic-program} that such a difference is of
little account: Herbrand bases are always extendible to meet
$\hb{P}=\hb{Q}$. The value of these observations is that by
implementing $\lpeq{v}$ we obtain an implementation for $\lpeq{}$ as
well. We will follow this strategy in Section
\ref{section:translation}.
Moreover, it is also clear by Proposition \ref{prop:weak-meets-visible}
that $\lpeq{v}$ is not a congruence for $\union$ and thus it does not
support program substitutions like $\lpeq{s}$.

Visible equivalence has its roots in the study of translation
functions \cite{Janhunen03:report,Janhunen06:jancl} and it was
proposed as a \emph{faithfulness} criterion for a translation function
$\trop{}$ between classes of programs, i.e., $P\lpeq{v}\tr{}{P}$
should hold for all programs $P$. The bijective relationship of stable
models ensures that a faithful translation (see Theorem
\ref{theorem:tr-sns-faithful} for an instance) preserves the number of
stable models. This is highly desirable in ASP where stable models
correspond to solutions of problems and the ability to count solutions
correctly after potential program transformations is of interest.
However, this is not guaranteed, if we consider weaker alternatives of
$\lpeq{v}$ obtained in a general framework based on {\em equivalence
frames} \cite{ETW05:ijcai}. Visible equivalence does not really fit
into equivalence frames based on {\em projected answer sets}. A
projective variant of Definition \ref{def:visible-equivalence} would
simply impose
%
$\sel{M\isect\hbv{P}}{M\in\sm{P}}=\sel{N\isect\hbv{Q}}{N\in\sm{Q}}$
%
on $P$ and $Q$ for $P\lpeq{vp}Q$ to hold, which is clearly implied by
$P\lpeq{v}Q$ but not vice versa. The key observation is that a weakly
faithful translation function $\trop{}$, i.e., $\trop{}$ satisfies
$P\lpeq{vp}\tr{}{P}$ for all $P$, does not necessarily preserve the
number of stable models --- contradicting the general nature of
ASP. As an illustration of these ideas, let us consider
$P=\set{a\IF\naf b\END b\IF\naf a\END}$
and
$Q_n=\tr{exp}{P}=P\union\sel{c_i\IF\naf d_i\END d_i\IF\naf c_i\END}{0<i\leq n}$
where $n>0$ is a parameter of $\trop{exp}$ and
$\hbv{Q}=\hbv{P}=\set{a,b}$ by definition. It follows that
$\sm{P}=\set{\set{a},\set{b}}$ and $Q_n$ has $2^{n+1}$ stable models
so that $M\isect\set{a,b}\in\sm{P}$ holds for each $M\in\sm{Q_n}$.
Therefore $P\lpeq{vp}Q_n$ but $P\not\lpeq{v}Q_n$ hold for every $n>0$,
i.e., $\trop{exp}$ would be faithful only in the weaker sense.
A drawback of translation functions like $\trop{exp}$ is that for
sufficiently large values of $n$, it is no longer feasible to count
the number of stable models of $P$ using its translation $Q_n$
which is only polynomially longer than $P$.

Equivalence relations play also a role in {\em forgetting}. Given a
logic program $P$ and a set of atoms $F\subseteq\hb{P}$, the goal is
to remove all instances of atoms of $F$ from $P$ but preserve the
semantics of $P$ as far as possible. \citeN{EW06:nmr} provide an
account of forgetting in the case of \emph{disjunctive} logic
programs. The result of forgetting $\forget{P}{F}$ is not
syntactically unique but its stable models are defined as the
$\subseteq$-minimal elements of
$\sm{P}\setminus F=\sel{M\setminus F}{M\in\sm{P}}$.
For instance, the program $P_n$ in Example \ref{ex:even-subsets}
is a valid result of forgetting if we remove
$F=\eset{\at{odd}_1}{\at{odd}_n}$ from $Q_n$.
We note that forgetting a set of atoms $F$ is somewhat analogous to
hiding $F$ in $P$, i.e., setting $\hbh{P}=F$, but obvious differences
are that $\hb{\forget{P}{F}}\isect F=\emptyset$ by definition and
forgetting can affect the number of stable models in contrast to
hiding.
Nevertheless, \citeN{EW06:nmr} show that forgetting preserves weak
equivalence in the sense that $P\lpeq{}Q$ implies
$\forget{P}{F}\lpeq{}\forget{Q}{F}$. This property is shared by
$\lpeq{v}$ in the fully visible case as addressed in Proposition
\ref{prop:weak-meets-visible}. In general, we can establish
the following.

\begin{proposition}
If $P\lpeq{v}Q$, then $\forget{P}{\hbh{P}}\lpeq{}\forget{Q}{\hbh{Q}}$.
\end{proposition}

\begin{proof}
Let us assume $P\lpeq{v}Q$ which implies both $\hbv{P}=\hbv{Q}$ and
the existence of a bijection $f$ in the sense of Definition
\ref{def:visible-equivalence}.
Assuming $\forget{P}{\hbh{P}}\not\lpeq{}\forget{Q}{\hbh{Q}}$, we
derive without loss of generality the existence of a stable model
$M\in\sm{\forget{P}{\hbh{P}}}$ such that
$M\not\in\sm{\forget{Q}{\hbh{Q}}}$. Note that $M$ is a subset of
$\hb{\forget{P}{\hbh{P}}}=\hbv{P}=\hbv{Q}=\hb{\forget{Q}{\hbh{Q}}}$
and a $\subseteq$-minimal element in $\sm{P}\setminus\hbh{P}$
defined in the preceding discussion.

Then consider any $M'\in\sm{P}$ such that $M=M'\setminus\hbh{P}$. It
follows by the properties of $f$ that $N'=f(M')\in\sm{Q}$ and
$N'\isect\hbv{Q}=M'\isect\hbv{P}=M$. Thus $M=N'\setminus\hbh{Q}$
belongs to $\sm{Q}\setminus\hbh{Q}$.
Let us then assume that $M$ is not $\subseteq$-minimal in this set,
i.e., there is $N\in\sm{Q}\setminus\hbh{Q}$ such that $N\subset
M$. Using the properties of $f$ and the same line of reasoning as
above for $M'$ and $N'$ but in the other direction, we learn that
$N\in\sm{P}\setminus\hbh{P}$ holds for $N\subset M$. A contradiction,
since $M$ is $\subseteq$-minimal in this set.

It follows that $M$ is also a $\subseteq$-minimal element in
$\sm{Q}\setminus\hbh{Q}$ so that $M\in\sm{\forget{Q}{\hbh{Q}}}$, a
contradiction.
\end{proof}

\subsection{%
Preliminary Analysis of the Equivalence Verification Problem}
\label{section:verification-problem}

The definition of stable models is based on the whole Herbrand base
$\hb{P}$ and hence it neglects which atoms are visible and which not.
The weak equivalence relation $\lpeq{}$ is based on the same line of
thinking and in \cite{JO02:jelia}, we presented methods for verifying
the weak equivalence of two programs $P$ and $Q$ satisfying
$\hb{P}=\hb{Q}$. A relatively naive approach is to cross-check the
stable models in $\sm{P}$ and $\sm{Q}$ in order to establish
$\sm{P}=\sm{Q}$ and thus $P\lpeq{}Q$.
The respective deterministic algorithm $\proc{EqNaive}$ is described
in Figure \ref{fig:naive} (a).%
\footnote{For the sake of brevity, compute statements
(\ref{eq:compute-statement}) are not covered by $\proc{EqNaive}$.}
The algorithm may use any algorithm such as the one given by
\citeN{SNS02:aij} for enumerating the stable models of $P$ and $Q$ one
at a time. Due to $\FNP$-completeness of the respective function
problem \cite{SNS02:aij}, the computation of each model may require
time exponential in the length of input, i.e., $\len{P}$ or $\len{Q}$.
The number of stable models to be cross-checked by a function
call $\proc{EqNaive}(P,Q)$ can also be exponential. However, the tests
for instability on lines 4 and 7 can be clearly accomplished in
polynomial time. This is because the least model $\lm{R}$ of any
positive set of rules $R$ can be computed in time linear in $\len{R}$
using a generalization of the procedure developed by \citeN{DG84:jlp}.

\begin{figure}
\figrule
\begin{center}
\begin{tabular}{ccc}
\begin{minipage}[t]{.4\textwidth}
\begin{tabbing}
xxx\=xx\=xx\=xx\=xx\=xx\=xx\=\kill
\ \,1:\> \kw{function} $\proc{EqNaive}(P,Q)$: Boolean; \\
\ \,2:\> \kw{begin} \\
\ \,3:\>\> \kw{for} $M\in\sm{P}$ \kw{do} \\
\ \,4:\>\>\> \kw{if} $M\neq\lm{\GLred{Q}{M}}$ \\
\ \,5:\>\>\>\> \kw{then return} false; \\
\ \,6:\>\> \kw{for} $N\in\sm{Q}$ \kw{do} \\
\ \,7:\>\>\> \kw{if} $N\neq\lm{\GLred{P}{N}}$ \\
\ \,8:\>\>\>\> \kw{then return} false; \\
\ \,9:\>\> \kw{return} true \\
10:\>\kw{end}
\end{tabbing}
\end{minipage}
&&
\begin{minipage}[t]{.4\textwidth}
\begin{tabbing}
xx\=xx\=xx\=xx\=xx\=xx\=xx\=\kill
1:\>\kw{algorithm} $\proc{NotEq}(P,Q)$; \\
2:\>\kw{begin} \\
3:\>\> \kw{choose} $M\subseteq\hb{P}$ \kw{and} $N\subseteq\hb{Q}$; \\
4:\>\> \kw{if} $M=\lm{\GLred{P}{M}}$ \kw{and} $M\neq\lm{\GLred{Q}{M}}$ \\
5:\>\>\> \kw{then accept}; \\
6:\>\> \kw{if} $N=\lm{\GLred{Q}{N}}$ \kw{and} $N\neq\lm{\GLred{P}{N}}$ \\
7:\>\>\> \kw{then accept}; \\
8:\>\> \kw{reject} \\
9:\>\kw{end}
\end{tabbing}
\end{minipage}
\\
(a) && (b)
\end{tabular}
\end{center}
\caption{A naive deterministic and a nondeterministic
algorithm for verifying $P\lpeq{}Q$ and $P\not\lpeq{}Q$,
respectively, when $\hb{P}=\hb{Q}$ and all atoms of $P$
and $Q$ are visible. \label{fig:naive}}
\figrule
\end{figure}

On the other hand, we get an upper limit for the computational time
complexity of the equivalence verification problem by inspecting the
nondeterministic algorithm presented in Figure \ref{fig:naive} (b).
The idea is to select an interpretation $M$ for $P$ (line 3) and to
verify that $M$ is a counter-example to $P\lpeq{}Q$ (lines 4--7). Both
tasks can be completed in time linear in $\len{P}+\len{Q}$. Since
$\proc{NotEq}(P,Q)$ accepts its input in the nondeterministic sense if
and only if $P\not\lpeq{}Q$, we see that the equivalence verifying
problem is a problem in $\coNP$.
On the other hand, checking the existence of a stable model for a
given logic program forms an $\NP$-complete decision problem
\cite{SNS02:aij}. Thus one can establish the $\coNP$-completeness of
the verification problem by reducing the complement of the latter
problem, i.e., checking that a logic program does {\em not} have
stable models, to the problem of verifying that $P$ is equivalent with
$\set{a\IF\naf a}$ --- a program having no stable models.
These observations on computational complexity suggest an alternative
computational strategy for solving the equivalence verification
problem \cite{JO02:jelia,OJ04:lpnmr}.  The idea is that
counter-examples for $P\lpeq{}Q$ are explicitly specified in terms of
rules and then proved non-existent using the same search algorithm as
what is used for the computation of stable models.

Unfortunately, further sources of complexity arise if we allow the use
of hidden atoms in \system{smodels} programs and consider $\lpeq{v}$ rather
than $\lpeq{}$. To see this, let us analyze how the operation of
$\proc{EqNaive}$ should be modified in order to deal with invisible
atoms. In fact, each cross-checking step has to be refined.  It is no
longer enough to compute a stable model $M$ for $P$. In addition to
this, we have to {\em count} how many stable models of $P$ coincide
with $M$ up to $\hbv{P}$, i.e., determine the number
$n=|\sel{N\in\sm{P}}{N\isect\hbv{P}=M\isect\hbv{P}}|$.
Then it is sufficient to check that $Q$ has equally many stable models
that coincide with $M$ up to $\hbv{P}$. This line of thinking applies
directly to the pair of programs given in Example
\ref{ex:visible-equivalence}.

By numbering stable models in the order they are encountered, we
obtain the basis for a bijective relationship as insisted by
Definition \ref{def:visible-equivalence}. The bad news is that the
computational complexity of counting models appears to be much higher
than finding a model; see \cite{Roth96:aij} for the case of
propositional satisfiability. Since classical models are easily
captured with stable models \cite{Niemela99:amai}, counting stable
models of a logic program cannot be easier than counting satisfying
assignments for a set of propositional clauses. Thus the complexity of
verifying $\lpeq{v}$ appears to be very high in general and
restrictions on visible atoms do not seem to provide us a way
circumvent the counting problem: If $\hbv{P}=\hbv{Q}=\emptyset$ is
assumed, then $P\lpeq{v}Q$ if and only if $P$ and $Q$ have the same
number of stable models.

In order to avoid model counting as discussed above, we should
restrict ourselves to logic programs $P$, for which the set
$\sel{N\in\sm{P}}{N\isect\hbv{P}=M\isect\hbv{P}}$
contains exactly one element for each $M\in\sm{P}$. Then stable models
$M,N\in\sm{P}$ can be distinguished in terms of visible atoms:
\begin{equation}
\label{eq:separable}
M\neq N \text{ implies } M\isect\hbv{P}\neq N\isect\hbv{P}.
\end{equation}

\begin{definition}
\label{def:separability}
Given an \system{smodels} program $P$, a set of interpretations
$C\subseteq\pset{\hb{P}}$ is {\em separable} with $\hbv{P}$ if
(\ref{eq:separable}) holds for all $M,N\in C$, and we say that $P$ has
separable stable models if $\sm{P}$ is separable with $\hbv{P}$.
\end{definition}

Unfortunately, the separability of $P$ and $Q$ does
not imply that $\proc{EqNaive}(P,Q)$ and $\proc{NotEq}(P,Q)$ work
correctly as $\hbh{P}$ and $\hbh{Q}$ differ and may lead to
unnecessary disqualification of models by the polynomial time tests
$M\neq\lm{\GLred{Q}{M}}$ and $N\neq\lm{\GLred{P}{N}}$.  These tests
capture correctly conditions $M\not\in\sm{Q}$ and $M\not\in\sm{P}$,
respectively, but when all atoms are visible.  However, a higher
computational complexity is involved in the presence of invisible
atoms.
E.g.\ the former test would have to be replaced by a computation
verifying that there is no $N\in\sm{Q}$ such that
$M\isect\hbv{P}=N\isect\hbv{Q}$ holds. This tends to push the worst
case time complexity of the equivalence verification problem to the
second level of polynomial time hierarchy
\cite{Stockmeyer76:tcs}.
Thus it seems that we need a stronger restriction than separability in
order to keep the problem of verifying $P\lpeq{v}Q$ as a decision
problem in $\coNP$ --- an obvious prerequisite for the
translation-based verification technique in \cite{JO02:jelia}.

\subsection{Programs Having Enough Visible Atoms}

In the fully visible case, the complexity of the verification problem
is alleviated by the computation of least models in algorithm
$\proc{NotEq}(P,Q)$. Those models are {\em unique} models associated
with the respective Gelfond-Lifschitz reductions of programs and they
provide the basis for detecting the (in)stability of model candidates.
Having such a unique model for each reduct is the key property that we
would like to carry over to the case of programs involving invisible
atoms. To achieve this, we propose a semantical restriction for the
class of logic programs as follows.
Given a logic program $P$ and a set of atoms $A\subseteq\hb{P}$, we
write $\vis{A}$ and $\hid{A}$ for $A\isect\hbv{P}$ and
$A\isect\hbh{P}$, respectively. Moreover, we are going to use
shorthands $A$, $B$, and $H$ for the respective sets of atoms
$\eset{a_1}{a_n}$, $\eset{b_1}{b_m}$, and $\eset{h_1}{h_l}$ appearing
in rules (\ref{eq:basic-rule}) -- (\ref{eq:compute-statement}).
Analogously, the notations $A=W_A$ and $\naf B=W_B$ capture the sets
of weights associated with $A$ and $B$ in the body of
(\ref{eq:weight-rule}).
The goal of Definition \ref{def:hidden-part} is to extract the
hidden part $\eval{\hid{P}}{\vis{I}}$ of an \system{smodels} program $P$
by partially evaluating it with respect to an interpretation
$\vis{I}\subseteq\hbv{P}$ for its visible part.

\begin{definition}
\label{def:hidden-part}
For an \system{smodels} program $P$ and an interpretation
$\vis{I}\subseteq\hbv{P}$ for the visible part of $P$, the hidden part
of $P$ relative $\vis{I}$, denoted $\eval{\hid{P}}{\vis{I}}$, contains

\begin{enumerate}
\item
a basic rule $h\IF\hid{A},\naf\hid{B}$
$\iff$
there is a basic rule $h\IF A,\naf B$ in $P$ such that
$h\in\hbh{P}$ and $\vis{I}\models\vis{A}\union\naf\vis{B}$;

\item
a choice rule $\choice{\hid{H}}\IF\hid{A},\naf\hid{B}$
$\iff$
there is a choice rule $\choice{H}\IF A,\naf B$ in $P$ such that
$H_h\neq\emptyset$ and $\vis{I}\models\vis{A}\union\naf\vis{B}$;

\item
a constraint rule $h\IF\cnt{c'}{\hid{A},\naf\hid{B}}$
$\iff$
there is a constraint rule $h\IF\cnt{c}{A,\naf B}$ in $P$ such that
$h\in\hbh{P}$ and
$c'=\max(0,c-|\sel{l\in\vis{A}\union\naf\vis{B}}{\vis{I}\models l}|)$;

\item
a weight rule $h\IF\limit{w'}{\hid{A}=W_{\hid{A}},\naf\hid{B}=W_{\hid{B}}}$
$\iff$
there is a weight rule $h\IF\limit{w}{A=W_A,\naf B=W_B}$ in $P$ such that
$h\in\hbh{P}$ and
\begin{equation}
\label{eq:visibly-reduced-weight-sum}
w'=\max(0,w-
   \wsum{\vis{I}}{\vis{A}=W_{\vis{A}},\naf\vis{B}=W_{\vis{B}}});
\end{equation}

\item
and no compute statements.
\end{enumerate}
\end{definition}

This construction can be viewed as a generalization of the
simplification operation $\simplify{P}{T}{F}$ proposed by Cholewinski
and Truszczy\'nski \citeyear{CT99:jlp} to the case of \system{smodels}
programs, but restricted in the sense that $T$ and $F$ are subsets of
$\hbv{P}$ rather than $\hb{P}$. More precisely put, we have
$\eval{\hid{P}}{\vis{I}}=\simplify{P}{\vis{I}}{\hbv{P}-\vis{I}}$
for a normal program, i.e., a set of basic rules $P$.

Roughly speaking, our idea is to allow the use of invisible atoms as
long as they do not interfere with the number of stable models
obtained for the visible part. We consider the invisible part of a
program ``well-behaving'' in this sense if and only if
$M=\lm{\GLred{(\eval{\hid{P}}{\vis{I}})}{M}}$
has a unique fixpoint $M$ for every $\vis{I}\subseteq\hbv{P}$.  In
particular, it should be pointed out that Definition
\ref{def:hidden-part} excludes compute statements which are not
supposed to affect this property (in perfect analogy to Definition
\ref{def:sm}).

\begin{definition}
\label{def:enough-visible-atoms}
An \system{smodels} program $P$ has enough visible atoms
if and only if $\eval{\hid{P}}{\vis{I}}$ has a unique stable
model for every $\vis{I}\subseteq\hbv{P}$.
\end{definition}

This property can be achieved for any program by making sufficiently
many atoms visible. To see this, consider Definition
\ref{def:hidden-part} when $\hbv{P}=\hb{P}$ and $\hbh{P}=\emptyset$:
it follows that $\eval{\hid{P}}{\vis{I}}=\emptyset$ for which the
existence of a unique stable model is immediate.
Generally speaking, verifying the property of having enough visible atoms
can be computationally quite hard in the worst case, but this property
favorably trades off the complexity of verifying $\lpeq{v}$ as we
shall see in Section \ref{section:translation}.

\begin{proposition}
\label{prop:eva-complexity}
Checking whether an \system{smodels} program $P$ has enough visible
atoms forms a $\coNP$-hard decision problem $\lang{EVA}$ that belongs
to $\coPH{2}$.
\end{proposition}

Here the language $\lang{EVA}$ consists of $\system{smodels}$ programs
that have enough visible atoms. For the proof, we introduce two
further languages $\lang{EVA}^{\leq 1}$ and $\lang{EVA}^{\geq 1}$.
The former includes those \system{smodels} programs $P$ for which
$\eval{\hid{P}}{\vis{I}}$ has {\em at most} one stable model for every
$\vis{I}\subseteq\hbv{P}$. The latter is defined analogously, but {\em
at least} one stable model for each $\eval{\hid{P}}{\vis{I}}$ is
demanded.  It should be now clear that
$\lang{EVA}=\lang{EVA}^{\leq 1}\isect\lang{EVA}^{\geq 1}$
which provides us a basis for complexity analysis.

\begin{proof}[Proof of Proposition \ref{prop:eva-complexity}]
To show that $\lang{EVA}_{\leq 1}\in\coNP$, we describe a
nondeterministic Turing machine $M_{>1}$ that accepts the complement of
$\lang{EVA}_{\leq 1}$.
Given a finite \system{smodels} program $P$ as input, the machine
$M_{>1}$ chooses nondeterministically two interpretations
$I,J\subseteq\hbv{P}$ and computes $Q=\eval{\hid{P}}{\vis{I}}$.  Then
$M_{>1}$ checks in polynomial time that $\vis{I}=\vis{J}$ and
$\hid{I}\neq\hid{J}$ as well as that both $\hid{I}$ and $\hid{J}$ are
stable models of $Q$.  If not, it rejects the input and accepts it
otherwise. It follows that $M_{>1}$ accepts $P$ in the
nondeterministic sense if and only if $P\not\in\lang{EVA}_{\leq
1}$. Hence $\lang{EVA}_{\leq 1}\in\coNP$.

The case of $\lang{EVA}_{\geq 1}$ is handled by presenting a
nondeterministic machine $M_{0}$ which uses an $\NP$ oracle and which
accepts the complement of $\lang{EVA}_{\geq 1}$. The machine $M_{0}$
chooses nondeterministically an interpretation
$\vis{I}\subseteq\hbv{P}$ for the input $P$. Then it computes
$\eval{\hid{P}}{\vis{I}}$ and consults an $\NP$-oracle
\cite{SNS02:aij} to check whether $\eval{\hid{P}}{\vis{I}}$ has a
stable model. If not, it accepts the input and rejects it
otherwise. Given the oracle, these computations can be accomplished in
polynomial time.  Now $M_0$ accepts $P$ in the nondeterministic sense
if and only if $P\not\in\lang{EVA}_{\geq 1}$. Thus we have established
that $\lang{EVA}_{\geq 1}\in\coPH{2}$.

We may now combine $M_{>1}$ and $M_0$ into one oracle machine $M$ that
accepts an \system{smodels} program $P$ $\iff$ $M_{>1}$ accepts $P$ or
$M_0$ accepts $P$. Equivalently, we have $P\not\in\lang{EVA}_{\leq 1}$
or $P\not\in\lang{EVA}_{\geq 1}$, i.e.,
$P\not\in(\lang{EVA}_{\leq 1}\isect\lang{EVA}_{\geq 1})=\lang{EVA}$.
Since $M$ is an oracle machine with an $\NP$ oracle, we have
actually shown that $\lang{EVA}\in\coPH{2}$.

To establish $\coNP$-hardness, we present a reduction from
$\lang{3SAT}$ to $\lang{EVA}$. So let us consider an instance of
$\lang{SAT}$, i.e., a finite set $S=\eset{C_1}{C_n}$ of three-literal
clauses $C_i$ of the form $l_1\lor l_2\lor l_3$ where each $l_i$ is
either an atom $a$ or its classical negation $\neg a$. Each clause $C_i$ is
translated into a rule $u\IF f_1,f_2,f_3$ where $f_i=a$ for $l_i=\neg
a$ and $f_i=\naf a$ for $l_i=a$. The outcome is an \system{smodels}
program $P_S$ which consists of clauses of $S$ translated in this way
plus two additional rules $s\IF\naf u$ and $x\IF s,\naf x$. Moreover,
we define $\hbv{P_S}=\hb{S}$ and $\hbh{P_S}=\set{u,s,x}$ so that
either $\eval{\hid{(P_S)}}{\vis{I}}=\set{u\END s\IF\naf u\END x\IF
s,\naf x}$ or $\eval{\hid{(P_S)}}{\vis{I}}=\set{s\IF\naf u\END x\IF
s,\naf x}$ depending on $\vis{I}\subseteq\hbv{P_S}$.
It follows that $S\in\lang{3SAT}$ $\iff$ there is an interpretation
$J\subseteq\hb{S}$ such that $J\models S$ $\iff$ there is an interpretation
$\vis{I}=J\subseteq\hbv{P_S}$ such that $u$ does not appear as a fact
in $\eval{\hid{(P_S)}}{\vis{I}}$ $\iff$ there is an interpretation
$\vis{I}\subseteq\hbv{P_S}$ such that $\eval{\hid{(P_S)}}{\vis{I}}$
has no stable models $\iff$ $P_S$ has not enough visible atoms,
since $\eval{\hid{(P_S)}}{\vis{I}}$ cannot have several stable models.
Thus we may conclude $\lang{EVA}$ to be $\coNP$-hard.
\end{proof}

Although the classification of $\lang{EVA}$ given in Proposition
\ref{prop:eva-complexity} is not exact, exponential worst case time
complexity should be clear.
However, there are certain syntactic classes of logic programs which
are guaranteed to have enough visible atoms and no computational
efforts are needed to verify this. For instance, programs $P$ for
which $\eval{\hid{P}}{\vis{I}}$ is always positive or {\em stratified}
\cite{ABW88} in some sense. Note that such syntactic restrictions need
not be imposed on the visible part of $P$ which may then fully
utilize expressiveness of rules.

\begin{example}
Consider logic programs
$P=\set{a\IF b\END}$,
$Q=\set{a\IF c\END c\IF b\END}$, and
$R=\set{a\IF\naf c\END c\IF\naf d\END d\IF b\END}$ with
$\hbv{P}=\hbv{Q}=\hbv{R}=\set{a,b}$.

Given $\vis{I}=\emptyset$, the hidden parts are
$\eval{\hid{P}}{\vis{I}}=\emptyset$,
$\eval{\hid{Q}}{\vis{I}}=\emptyset$, and
$\eval{\hid{R}}{\vis{I}}=\set{c\IF\naf d\END}$ for which
we obtain unique stable models
$M_P=M_Q=\emptyset$ and $M_R=\set{c}$.
On the other hand, we obtain $\eval{\hid{P}}{\vis{J}}=\emptyset$,
$\eval{\hid{Q}}{\vis{J}}=\set{c\END}$, and
$\eval{\hid{R}}{\vis{J}}=\set{c\IF\naf d\END d\END}$ for
$\vis{J}=\set{a,b}$. Thus the respective unique stable models of the
hidden parts are $N_P=\emptyset$ and $N_Q=\set{c}$, and $N_R=\set{d}$.
\end{example}

Next we relate the property of having enough visible atoms with the
model separation property.  The proof of Lemma \ref{lemma:uniqueness}
takes place in \ref{appendix:proofs}.

\begin{lemma}
\label{lemma:uniqueness}
Let $P$ be an \system{smodels} program.
If $M\subseteq\hb{P}$ is a stable model of $P$, then $\hid{M}$ is a
stable model of $\eval{\hid{P}}{\vis{M}}$ as given in Definition
\ref{def:hidden-part}.
\end{lemma}

\begin{proposition}
\label{prop:separability-and-visibility}
Let $P$ be an \system{smodels} program.
If $P$ has enough visible atoms, then $P$ has separable stable models.
\end{proposition}

\begin{proof}
Suppose that $P$ is an \system{smodels} program which has enough
visible atoms but $\sm{P}$ is not separable with $\hbv{P}$.
Then there are two stable models $N,M\in\sm{P}$ such that
$\vis{M}=\vis{N}$ but $\hid{M}\neq\hid{N}$. Thus $\hid{M}$ and
$\hid{N}$ are stable models of
$\eval{\hid{P}}{\vis{M}}=\eval{\hid{P}}{\vis{N}}$ by Lemma
\ref{lemma:uniqueness}.  A contradiction as $P$ has enough visible atoms.
\end{proof}

The converse of Proposition \ref{prop:separability-and-visibility}
does not hold in general.  Consider, for instance $P=\set{a\IF\naf
a\END b\IF a,\naf b\END}$ with $\hbv{P}=\set{a}$. Since
$\sm{P}=\emptyset$, it is trivially separable with $\hbv{P}$.  But for
$\vis{I}=\set{a}\subseteq\hbv{P}$, the hidden part
$\eval{\hid{P}}{\vis{I}}=\set{b\IF\naf b\END}$ has no stable models
and thus $P$ does not have enough visible atoms.


\section{Translation-Based Verification}
\label{section:translation}

In this section, we concentrate on developing a translation-based
verification technique for visible equivalence, i.e., the relation
$\lpeq{v}$ introduced in Section \ref{section:equivalence}. Roughly
speaking, our idea is to translate given \system{smodels} programs $P$
and $Q$ into a single \system{smodels} program $\eqt(P,Q)$ which has a
stable model if and only if $P$ has a stable model $M$ for which $Q$
does {\em not} have a stable model $N$ such that $\vis{M}=\vis{N}$.
We aim to use such a translation for finding a {\em counter-example}
for $P\lpeq{v}Q$ when $\hbv{P}=\hbv{Q}$ and both $P$ and $Q$ have
enough visible atoms.
Note that if $\hbv{P}\neq\hbv{Q}$, then $P\not\lpeq{v}Q$ follows
directly by Definition \ref{def:visible-equivalence}. As already
discussed in Section \ref{section:equivalence}, we need the property
of having enough visible atoms to trade off computational complexity
so that a polynomial translation is achievable. The good news is that
the programs produced by the front-end \system{lparse} have this
property by default unless too many atoms are explicitly hidden by the
programmer.
Our strategy for finding a counter-examples $M$ is
based on the following four steps.

\begin{enumerate}
\item
Find a stable model $M\in\sm{P}$ for $P$.

\item
Find the unique stable model $\hid{N}$ for $\eval{\hid{Q}}{\vis{M}}$.

\item
Form an interpretation $N=\vis{M}\union\hid{N}$.

\item
Check that $N\not\in\sm{Q}$, i.e., $N\neq\lm{\GLred{Q}{N}}$ or
$N\not\models\compst{Q}$.
\end{enumerate}

Here the idea is that the uniqueness of $\hid{N}$ with respect to
$\vis{M}$ excludes the possibility that $Q$ could possess another
stable model $N'\neq N$ such that $\vis{M}=\vis{N'}$. This follows
essentially by Lemma \ref{lemma:uniqueness}: if $N'\in\sm{Q}$ were the
case, then $\hid{N'}$ would be unique for $\vis{N'}=\vis{M}=\vis{N}$,
i.e., $\hid{N'}=\hid{N}$ and $N=N'$.

In the sequel, we present a translation function $\eqt$ that
effectively captures the four steps listed above within a single
\system{smodels} program $\eqt(P,Q)$. In order to combine several
programs in one, we have to rename atoms and thus introduce {\em new}
atoms outside $\hb{P}\union\hb{Q}$:
\begin{itemize}
\item
a new atom $\renh{a}$ for each atom $a\in\hbh{Q}$ and

\item
a new atom $\ren{a}$ for each atom $a\in\hb{Q}$.
\end{itemize}
The former atoms will be used in the representation of
$\eval{\hid{Q}}{\vis{M}}$ while the latter are to appear in the
translation of $\GLred{Q}{N}$. The intuitive readings of $\renh{a}$
and $\ren{a}$ are that $a\in\hid{N}$ and $a\in\lm{\GLred{Q}{N}}$ hold,
respectively.
For notational convenience, we extend the notations $\renh{a}$ and
$\ren{a}$ for sets of atoms $A$ as well as sets of positive rules $R$
in the obvious way. For instance, $\renh{A}$ denotes
$\sel{\renh{a}}{a\in A}$ for any $A\subseteq\hbh{Q}$.

\begin{definition}
\label{def:translation}
Let $P$ and $Q$ be \system{smodels} programs such that $\hbv{P}=\hbv{Q}$.
The translation
$\eqt(P,Q)=P\union\trhid{Q}\union\trlm{Q}\union\unstable{Q}$
extends $P$ with three sets of rules to be made precise by
Definitions \ref{def:translation-hidden}--\ref{def:translation-unstable}.
Atoms $c$, $d$, and $e$ introduced in Definition
\ref{def:translation-unstable} are assumed to be new.
\end{definition}

As regards our strategy for representing counter-examples, the rules
in the translation $\eqt(P,Q)$ play the following roles. The rules of
$P$ capture a stable model $M$ for $P$ while the rest of the
translation ensures that $Q$ does not have a stable model $N$ such
that $\vis{M}=\vis{N}$.
To make the forthcoming definitions more accessible for the reader, we
use simple normal programs
$P=\set{a\IF\naf b\END\ b\IF\naf a\END}$
and
$Q=\set{a\IF b,\naf a\END\ b\IF\naf a\END}$
with $\hbv{P}=\hbv{Q}=\set{a}$ as our running example. The rules
contributed by Definitions
\ref{def:translation}--\ref{def:translation-unstable}
are collected together in Fig.~\ref{fig:eqt}.

\begin{definition}
\label{def:translation-hidden}
The translation $\trhid{Q}$ of an \system{smodels} program $Q$ contains
\begin{enumerate}
\item \label{item:Q-h-begin}
a basic rule
$\renh{h}\IF\renh{\hid{A}},\vis{A},\naf\renh{\hid{B}},\naf\vis{B}$
for each basic rule \\ $h\IF A,\naf B$ in $Q$ with $h\in\hbh{Q}$;

\item
a constraint rule
$\renh{h}\IF\cnt{c}{\renh{\hid{A}},\vis{A},\naf\renh{\hid{B}},\naf\vis{B}}$
for each constraint rule \\
$h\IF\cnt{c}{A,\naf B}$ in $Q$ with $h\in\hbh{Q}$;

\item
a choice rule
$\choice{\renh{\hid{H}}}\IF\renh{\hid{A}},\vis{A},
                           \naf\renh{\hid{B}},\naf\vis{B}$
for each choice rule \\ $\choice{H}\IF A,\naf B$ in $Q$ with
$\hid{H}\neq\emptyset$; and

\item \label{item:Q-h-end}
a weight rule
$\renh{h}\IF\limit{w}{\renh{\hid{A}}=W_{\renh{\hid{A}}},
                      \vis{A}=W_{\vis{A}},
                      \naf\renh{\hid{B}}=W_{\renh{\hid{B}}},
                      \naf\vis{B}=W_{\vis{B}}}$
for each weight rule $h\IF\limit{w}{A=W_A,\naf B=W_B}$ in $Q$ with
$h\in\hbh{Q}$.
\end{enumerate}
\end{definition}

The translation $\trhid{Q}$ includes rules which provide a
representation for the hidden part $\eval{\hid{Q}}{\vis{M}}$ which
depends dynamically on $\vis{M}$. This is achieved by leaving the
visible atoms from $\hbv{Q}=\hbv{P}$ untouched. However, the hidden
parts of rules are renamed systematically using atoms from
$\renh{\hbh{Q}}$.  This is to capture the unique stable model
$\hid{N}$ of $\eval{\hid{Q}}{\vis{M}}$ but renamed as
$\renh{\hid{N}}$.
\footnote{For the sake of simplicity, it is assumed that $Q$ does
not involve compute statements referring to invisible atoms in
order to achieve the property of having enough visible atoms.}
In our running example, the program $Q$ has only one rule with a
hidden atom $b$ in its head and that rule gets translated into
$\renh{b}\IF\naf a$ due to the visibility of $a$.

\begin{definition}
\label{def:translation-lm}
The translation $\trlm{Q}$ of an \system{smodels} program $Q$ consists of
\begin{enumerate}
\item \label{item:Q-begin}
a rule $\ren{h}\IF\ren{A},\naf\vis{B},\naf\renh{\hid{B}}$
for each basic rule $h\IF A,\naf B$ in $Q$;

\item
a rule
$\ren{h}\IF\cnt{c}{\ren{A},\naf\vis{B},\naf\renh{\hid{B}}}$
for each constraint rule $h\IF\cnt{c}{A,\naf B}$ in $Q$;

\item \label{item:translate-choice}
a rule
$\ren{h}\IF \ren{A}\union\set{h},\naf\vis{B},\naf\renh{\hid{B}}$
(resp.\ 
 $\ren{h}\IF \ren{A}\union\set{\renh{h}},\naf\vis{B},\naf\renh{\hid{B}}$)
for each choice rule $\choice{H}\IF A,\naf B$ in $Q$ and head
atom $h\in\vis{H}$ (resp. $h\in\hid{H}$); and

\item \label{item:Q-end}
a rule
$\ren{h}\IF\limit{w}{\ren{A}=W_{\ren{A}},\naf\vis{B}=W_{\vis{B}},
                     \naf\renh{\hid{B}}=W_{\renh{\hid{B}}}}$
for each weight rule \\ $h\IF\limit{w}{A=W_A,\naf B=W_B}$ in $Q$.
\end{enumerate}
\end{definition}

The rules in $\trlm{Q}$ catch the least model $\lm{\GLred{Q}{N}}$ for
$N=\vis{M}\union\hid{N}$ but expressed in $\ren{\hb{Q}}$ rather than
$\hb{Q}$. Note that $N$ is represented as
$\vis{M}\union\renh{\hid{N}}$ which explains the treatment of negative
body literals on the basis of visibility in these rules.
Two rules result for our running example. The negative literal $\naf
a$ appearing in the bodies of both rules is not subject to renaming
because $a$ is visible.

\begin{definition}
\label{def:translation-unstable}
The translation $\unstable{Q}$ of an \system{smodels} program $Q$ includes
\begin{enumerate}
\item \label{item:diff1}
rules $d\IF a, \naf\ren{a}$ and $d\IF \ren{a}, \naf a$
for each $a\in\hbv{Q}$;

\item \label{item:diff2}
rules $d\IF\renh{a},\naf\ren{a}$ and $d\IF\ren{a},\naf\renh{a}$
for each $a\in\hbh{Q}$;

\item \label{item:positive-consistency}
a rule $c\IF\naf\ren{a},\naf d$ for each positive literal $a\in\compst{Q}$;

\item \label{item:negative-consistency}
a rule $c\IF\ren{b},\naf d$ for each negative literal $\naf b\in\compst{Q}$;

\item \label{item:summarize}
rules $e\IF c$ and $e\IF d$; and

\item \label{item:require}
a compute statement $\compute{e}$.
\end{enumerate}
\end{definition}

The purpose of $\unstable{Q}$ is to disqualify $N$ as a stable model
of $Q$. The rules in Items \ref{item:diff1} and \ref{item:diff2} check
if $N$ and $\lm{\GLred{Q}{N}}$ differ. If not, then the rules in Items
\ref{item:positive-consistency} and \ref{item:negative-consistency}
check if $\lm{\GLred{Q}{N}}$ violates some compute statement of
$Q$. The rules in Item \ref{item:summarize} summarize the two possible
reasons why $Q$ does not have a stable model $N$ such that
$\vis{M}=\vis{N}$. This is then insisted by the compute statement in
Item \ref{item:require}.
In our running example, the program $Q$ is free of compute statements
and hence only rules for $d$ and $e$ are included in the translation.

\begin{figure}[t]
\figrule
\begin{tabular}{rl}
$P$: &
$a\IF\naf b\END$~~$b\IF\naf a\END$ \\
$\trhid{Q}:$ &
$\renh{b}\IF\naf a\END$ \\
$\trlm{Q}:$ &
$\ren{a}\IF\ren{b},\naf a\END$~~$\ren{b}\IF\naf a\END$\\
$\unstable{Q}:$ &
$d\IF a, \naf\ren{a}\END$ $d\IF \ren{a}, \naf a\END$ \\
& $d\IF\renh{b},\naf\ren{b}\END$ $d\IF\ren{b},\naf\renh{b}\END$ \\
& $e\IF c\END$ $e\IF d\END$ \\
& $\compute{e}\END$
\end{tabular}
\caption{The rules of the translation $\eqt(P,Q)$ for
         $P=\set{a\IF\naf b\END b\IF\naf a\END}$ and
         $Q=\set{a\IF b,\naf a\END b\IF\naf a\END}$ where
         $a$ is visible and $b$ is hidden.
         \label{fig:eqt}}
\figrule
\end{figure}

\begin{example}
The translation $\eqt(P,Q)$ given in Fig.~\ref{fig:eqt} has
two stable models
$\set{a,d,e}$
and
$\set{b,\renh{b},\ren{a},\ren{b},d,e}$
from which we can read off counter-examples
$M_1=\set{a}$ and $M_2=\set{b}$ for $P\lpeq{v}Q$ and the respective
disqualified interpretations for $Q$, i.e.,
$N_1=\set{a}$ and $N_2=\set{b}$.
The models $\lm{\GLred{Q}{N_1}}=\emptyset$ and
$\lm{\GLred{Q}{N_2}}=\set{a,b}$ are also easy to extract by projecting
the stable models of $\eqt(P,Q)$ with $\set{\ren{a},\ren{b}}$.
\end{example}

As regards the translation $\eqt(P,Q)$ as whole, we note that
$\hb{\eqt(P,Q)}$ equals to
$\hb{P}\union\renh{\hbh{Q}}\union\ren{\hb{Q}}\union\set{c,d,e}$.
Moreover, the translation is close to being linear. Item
\ref{item:translate-choice} in Definition \ref{def:translation} makes
an exception in this respect, but linearity can be achieved in
practise by introducing a new atom $b_r$ for each choice rule $r$.
Then the rules in the fourth item can be replaced by
$\ren{h}\IF h,b_r$ (resp. $\ren{h}\IF\renh{h},b_r$)
and $b_r\IF\ren{A},\naf\vis{B},\naf\renh{\hid{B}}$.
However, we use the current definition in order to avoid
the introduction of further new atoms.

Let us then address the correctness of the translation $\eqt(P,Q)$.
We begin by computing the Gelfond-Lifschitz reduct of the translation
$\eqt(P,Q)$.

\newcounter{myctr}

\begin{lemma}
\label{lemma:reduct}
Let $P$ and $Q$ be two \system{smodels} programs such that
$\hbv{P}=\hbv{Q}$ and
$I\subseteq\hb{P}\union\renh{\hbh{Q}}\union\ren{\hb{Q}}\union\set{c,d,e}$
an interpretation of $\eqt(P,Q)$.
Moreover, define $M=I\isect\hb{P}$,
$\hid{N}=\sel{a\in\hbh{Q}}{\renh{a}\in I}$, $N=\vis{M}\union\hid{N}$,
and $L=\sel{a\in\hb{Q}}{\ren{a}\in I}$ so that
$\renh{\hid{N}}=I\isect\renh{\hbh{Q}}$ and
$\ren{L}=I\isect\ren{\hb{Q}}$.

The Gelfond-Lifschitz reduct $\GLred{\eqt(P,Q)}{I}$ consists of
$\GLred{P}{M}$ extended by reducts $\GLred{\trhid{Q}}{I}$,
$\GLred{\trlm{Q}}{I}$, and $\GLred{\unstable{Q}}{I}$ specified as follows.

First, the reduct $\GLred{\trhid{Q}}{I}$ includes
\begin{enumerate}
\item \label{item:reduced-Q-h-begin}
a rule
$\renh{h}\IF\renh{\hid{A}},\vis{A}$
$\iff$
there is a basic rule $h\IF A,\naf B$ in $Q$ such that $h\in\hbh{Q}$,
and $N\models\naf B$;

\item
a rule
$\renh{h}\IF\cnt{c'}{\renh{\hid{A}},\vis{A}}$ where
$c'=\max(0,c-|\sel{b\in B}{N\models\naf b}|)$
$\iff$
there is a constraint rule $h\IF\cnt{c}{A,\naf B}$ in $Q$ such that
$h\in\hbh{Q}$;

\item
a rule
$\renh{h}\IF\renh{\hid{A}},\vis{A}$
$\iff$
there is a choice rule $\choice{H}\IF A,\naf B$ in $Q$ such that
$h\in\hid{H}\neq\emptyset$, $\hid{N}\models h$, and
$N\models\naf B$; and

\item \label{item:reduced-Q-h-end}
a rule
$\renh{h}\IF\limit{w'}{\renh{\hid{A}}=W_{\renh{\hid{A}}},\vis{A}=W_{\vis{A}}}$
where $w'=\max(0,w-\wsum{N}{\naf B=W_B})$
$\iff$
there is a weight rule $h\IF\limit{w}{A=W_A,\naf B=W_B}$ in $Q$ such
that $h\in\hbh{Q}$.
\setcounter{myctr}{\value{enumi}}
\end{enumerate}

Second, the reduct $\GLred{\trlm{Q}}{I}$ consists of
\begin{enumerate}
\setcounter{enumi}{\value{myctr}}
\item \label{item:reduced-Q-begin}
a rule $\ren{h}\IF\ren{A}$ $\iff$
there is a basic rule $h\IF A,\naf B$ in $Q$ such that
$N\models\naf B$;

\item
a rule $\ren{h}\IF\cnt{c'}{\ren{A}}$ where
$c'=\max(0,c-|\sel{b\in B}{N\models\naf b}|)$
$\iff$
there is a constraint rule $h\IF\cnt{c}{A,\naf B}$ in $Q$;

\item
a rule $\ren{h}\IF\ren{A}\union\set{h}$
(resp.\ $\ren{h}\IF\ren{A}\union\set{\renh{h}}$)
$\iff$
there is a choice rule $\choice{H}\IF A,\naf B$ in $Q$ with
$h\in\vis{H}$ (resp. $h\in\hid{H}$) such that $N\models\naf B$; and

\item \label{item:reduced-Q-end}
a rule
$\ren{h}\IF\limit{w'}{\ren{A}=W_{\ren{A}}}$ where
$w'=\max(0,w-\wsum{N}{\naf B=W_B})$
$\iff$
there is a weight rule $h\IF\limit{w}{A=W_A,\naf B=W_B}$ in $Q$.
\setcounter{myctr}{\value{enumi}}
\end{enumerate}

Third, the set $\GLred{\unstable{Q}}{I}$ contains
\begin{enumerate}
\setcounter{enumi}{\value{myctr}}
\item \label{item:reduced-rest-begin}
a rule $d\IF a$
$\iff$
there is $a\in\hbv{Q}$ such that $L\not\models a$;

\item
a rule $d\IF\renh{a}$
$\iff$
there is $a\in\hbh{Q}$ such that $L\not\models a$;

\item
a rule $d\IF\ren{a}$
$\iff$
there is $a\in\hb{Q}$ such that $N\not\models a$;

\item
the fact $c\IF$
$\iff$
there is $a\in\compst{Q}$ such that $L\not\models a$ and $I\not\models d$;

\item
a rule $c\IF\ren{b}$
$\iff$
there is $\naf b\in\compst{Q}$ and $I\not\models d$; and

\item \label{item:reduced-rest-end}
the rules $e\IF c$ and $e\IF d$.
\end{enumerate}
\end{lemma}

Lemma \ref{lemma:reduct} can be easily verified by inspecting the
definition of the translation $\eqt(P,Q)$ (Definitions
\ref{def:translation}--\ref{def:translation-unstable}) rule by rule
and using the definitions of $M$, $N$, and $L$ as well as the
generalization of Gelfond-Lifschitz reduct for the various rule types
(Definition \ref{def:reduct}).
The following proposition summarizes a number properties of
$\lm{\GLred{\eqt(P,Q)}{I}}$ which are used in the sequel to prove our
main theorem.

\begin{proposition}
\label{prop:least-model-of-reduct}
Let $P$, $Q$, $I$, $M$, $N$, and $L$ be defined
as in Lemma \ref{lemma:reduct}.
Define conditions (i) $M=\lm{\GLred{P}{M}}$, (ii)
$\hid{N}=\lm{\GLred{(\eval{\hid{Q}}{\vis{M}})}{\hid{N}}}$, and (iii)
$L=\lm{\GLred{Q}{N}}$.
\begin{enumerate}
\item \label{item:least-model-of-reduct1}
$\lm{\GLred{\eqt(P,Q)}{I}}\isect\hb{P}=\lm{\GLred{P}{M}}$.

\item \label{item:least-model-of-reduct2}
If (i), then
$\lm{\GLred{\eqt(P,Q)}{I}}\isect\renh{\hbh{Q}}=
 \renh{\lm{\GLred{(\eval{\hid{Q}}{\vis{M}})}{\hid{N}}}}$.

\item \label{item:least-model-of-reduct3}
If (i) and (ii), then
$\lm{\GLred{\eqt(P,Q)}{I}}\isect\ren{\hb{Q}}=\ren{\lm{\GLred{Q}{N}}}$.

\item \label{item:least-model-of-reduct4}
If (i), (ii), and (iii), then the set of atoms
$A=\lm{\GLred{\eqt(P,Q)}{I}}\isect\set{c,d,e}$ satisfies
\begin{enumerate}
\item
$d\in A$ $\iff$ $N\neq L$,
\item
$c\in A$ $\iff$ $d\not\in I$ and $L\not\models\compst{Q}$, and
\item
$e\in A$ $\iff$ $c\in A$ or $d\in A$.
\end{enumerate}
\end{enumerate}
\end{proposition}

\begin{theorem}
\label{theorem:correctness}
Let $P$ and $Q$ be two \system{smodels} programs such that $\hbv{P}=\hbv{Q}$
and $Q$ has enough visible atoms.
Then the translation $\eqt(P,Q)$ has a stable model if and only if
there is $M\in\sm{P}$ such that for all $N\in\sm{Q}$, $\vis{N}\neq\vis{M}$.
\end{theorem}

The proofs of Proposition \ref{prop:least-model-of-reduct} and Theorem
\ref{theorem:correctness} are given in \ref{appendix:proofs}. As a
corollary of Theorem \ref{theorem:correctness}, we obtain a new method
for verifying the visible equivalence of $P$ and $Q$. Weak equivalence
$\lpeq{}$ is covered by making all atoms of $P$ and $Q$ visible which
implies that the programs in question have enough visible atoms.

\begin{corollary}
\label{corollary:test-equivalence}
Let $P$ and $Q$ be two \system{smodels} programs so that
$\hbv{P}=\hbv{Q}$ and both $P$ and $Q$ have enough visible atoms.
Then $P\lpeq{v}Q$ if and only if the translations
$\eqt(P,Q)$ and $\eqt(Q,P)$ have no stable models.
\end{corollary}

\subsection{Computational Complexity Revisited}
\label{section:complexity}

In this section, we review the computational complexity of verifying
visible equivalence of \system{smodels} programs using the reduction
involved in Theorem \ref {theorem:correctness}.
First, we will introduce languages corresponding to the decision
problems of our interest and analyze their worst-case time
complexities.  The main goal is to establish that the verification of
visible equivalence forms a $\coNP$-complete decision problem for
\system{smodels} programs that have enough visible atoms.

\begin{definition}
\label{def:decision-problems}
For any \system{smodels} programs $P$ and $Q$,
\begin{itemize}
\item
$P\in\lang{SM}$ $\iff$ there is a stable model $M\in\sm{P}$;

\item
$\pair{P}{Q}\in\lang{IMPR}$ $\iff$ $\hbv{P}=\hbv{Q}$ and for each
$M\in\sm{P}$, there is $N\in\sm{Q}$ such that $\vis{N}=\vis{M}$;

\item
$\pair{P}{Q}\in\lang{IMPL}$ $\iff$ $\pair{Q}{P}\in\lang{IMPR}$; and

\item
$\pair{P}{Q}\in\lang{EQV}$ $\iff$ $P\lpeq{v} Q$. 
\end{itemize}
\end{definition}

The computational complexity of $\lang{SM}$ is already
well-understood: it forms an $\NP$-complete decision problem
\cite{MT91:jacm,SNS02:aij} and thus its complement
$\lang{\compl{SM}}$ is $\coNP$-complete.

\begin{theorem}
\label{theorem:IMPR-and-IMPL}
$\lang{IMPR}$, $\lang{IMPL}$, and $\lang{EQV}$ are $\coNP$-complete
decision problems for \system{smodels} programs having enough visible
atoms.
\end{theorem}

\begin{proof}
Let us establish that (i) $\lang{IMPR}\in\coNP$ and (ii)
$\lang{\compl{SM}}$ can be reduced to $\lang{IMPR}$.
\begin{enumerate}
\item[(i)]
Let $P$ and $Q$ be two \system{smodels} programs having enough visible
atoms.  Then define a reduction $r$ from $\lang{IMPL}$ to
$\lang{\compl{SM}}$ by setting $r(P,Q)=\eqt(P,Q)$ if $\hbv{P}=\hbv{Q}$
and $r(P,Q)=\emptyset$ otherwise. To justify that $r(P,Q)$ can be
computed in polynomial time we make the following observations.
The condition $\hbv{P}=\hbv{Q}$ can be verified in linear time if
atoms in $\hbv{P}$ and $\hbv{Q}$ are ordered, e.g., alphabetically.
If not, sorting can be done in time of $\order{n\log n}$ where
$n=\max(|\hbv{P}|,|\hbv{Q}|)$. Moreover, we can identify four
subprograms of $\eqt(P,Q)$, i.e., $P$, $\trhid{Q}$, $\trlm{Q}$, and
$\unstable{Q}$ in Definition \ref{def:translation} whose lengths
depend mostly linearly on $\len{P}$, $\len{Q}$, and $|\hb{Q}|$,
respectively. The rules of Item \ref{item:translate-choice} make
the only exception with a quadratic blow-up.

It follows by Definition \ref{def:decision-problems} and Theorem
\ref{theorem:correctness} that $\pair{P}{Q}\in\lang{IMPR}$ $\iff$
$r(P,Q)\not\in\lang{SM}$, i.e., $r(P,Q)\in\lang{\compl{SM}}$.  Since
$\lang{\compl{SM}}\in\coNP$ \cite{SNS02:aij} and $r$ is a polynomial time
reduction from $\lang{IMPR}$ to $\lang{\compl{SM}}$,
$\lang{IMPR}\in\coNP$.

\item [(ii)] 
Let $R$ be any \system{smodels} program. Now $R\in\lang{\compl{SM}}$
$\iff$ $R\not\in\lang{SM}$ $\iff$ $\sm{R}=\emptyset$.
Then consider any \system{smodels} program $\bot$ having no stable
models, i.e., $\sm{\bot}=\emptyset$, with a visible Herbrand base
$\hbv{\bot}=\hbv{R}$.  It follows that $\sm{R}=\emptyset$ $\iff$
$\pair{R}{\bot}\in\lang{IMPR}$. Thus $R\in\lang{\compl{SM}}$ $\iff$
$\pair{R}{\bot}\in\lang{IMPR}$.
\end{enumerate}

Items (i) and (ii) above imply that $\lang{IMPR}$ is $\coNP$-complete.
The classification of $\lang{IMPL}$ follows by a trivial reduction
$\pair{P}{Q}\in\lang{IMPR}$ $\iff$ $\pair{Q}{P}\in\lang{IMPL}$ that
works in both directions, i.e., from $\lang{IMPR}$ to $\lang{IMPL}$
and back.

The case of $\lang{EQV}$ follows. Definitions
\ref{def:visible-equivalence} and \ref{def:decision-problems} imply
that $\pair{P}{Q}\in\lang{EQV}$ $\iff$ $\pair{P}{Q}\in\lang{IMPR}$ and
$\pair{P}{Q}\in\lang{IMPL}$. Thus
$\lang{EQV}=\lang{IMPR}\cap\lang{IMPL}$ and $\lang{EQV}\in\coNP$ as
$\coNP$ is closed under intersection. The $\coNP$-hardness of
$\lang{EQV}$ follows easily as it holds for any \system{smodels}
program $R$ and a trivial \system{smodels} program $\bot$ with
$\sm{\bot}=\emptyset$ and $\hbv{\bot}=\hbv{R}$ that
$R\in\lang{\compl{SM}}$ $\iff$ $\pair{R}{\bot}\in\lang{EQV}$. Thus we
may conclude that $\lang{EQV}$ is $\coNP$-complete.
\end{proof}


\section{Weight Constraint Programs}
\label{section:weightconstraint}

The verification method presented in Section \ref{section:translation}
covers the class of \system{smodels} programs as defined in Section
\ref{section:syntax}. This class corresponds very closely to the input
language of the \system{smodels} search engine but it excludes {\em
optimization statements} which will not be addressed in this paper.
In this section we concentrate on extending our translation-based
verification method for the input language supported by the front-end
of the \system{smodels} system, namely \system{lparse}
\cite{LPARSE01:manual,SN01:lpnmr}.  The class of {\em weight
constraint programs} \cite{SNS02:aij} provides a suitable abstraction
of this language in the propositional case.
\footnote{Since \system{lparse} is responsible for instantiating
variables and pre-interpreting certain function symbols the input
language is actually much more general.}
\citeN{SNS02:aij} show how weight constraint programs can be
transformed into \system{smodels} programs using a modular translation
that introduces new atoms. Our strategy is to use this translation
to establish that the weak equivalence of two weight constraint
programs reduces to the visible equivalence of the respective
translations.

Next we introduce the syntax and semantics of weight constraint
programs. Recalling the syntax of weight rules (\ref{eq:weight-rule}),
a natural way to extend their expressiveness is to allow more
versatile use of weights as well as constraints associated with
them. This is achieved by recognizing weight constraints as
first-class citizens and using them as basic building blocks of rules
instead of plain atoms.

\begin{definition}
\label{def:wc}
A weight constraint $C$ is an expression of the form
\begin{equation}
\label{eq:wc}
l\leq
\{a_1=w_{a_1},\ldots,a_n=w_{a_n},
  \naf b_1=w_{b_1},\ldots,\naf b_m=w_{b_m}\}
\leq u, 
\end{equation}
where $a_i$'s and $b_j$'s are atoms, and $l$, $u$,
$w_{a_i}$'s, and $w_{b_j}$'s are natural numbers.
\end{definition}

As before, we use a shorthand $l\leq\set{A=W_{A},\naf B=W_B}\leq u$ for
a weight constraint (\ref{eq:wc}) where $A=\eset{a_1}{a_n}$ and
$B=\eset{b_1}{b_m}$ are the sets of atoms appearing in the constraint.
The numbers $l$ and $u$ give the respective \emph{lower and upper bounds}
for the constraint.  Definition \ref{def:wc} can be extended to the case
where integers rather than natural numbers are used as
weights. However, negative weights can be translated away
\cite{SNS02:aij} from weight constraints.

\begin{definition}
\label{def:wc-rule}
A weight constraint rule is an expression of the form
\begin{eqnarray}
\label{eq:wc-rule}
C_0\IF C_1,\ldots,C_r
\label{wc-rule}
\end{eqnarray}
where $C_i$ is a weight constraint for each $i\in\eset{0}{r}$.
\end{definition}

A weight constraint program is a program consisting of weight
constraint rules. As a weight constraint rule (\ref{wc-rule}) is a
generalization of a weight rule (\ref{eq:weight-rule}), we can define
the satisfaction relation for weight constraint programs in analogy to
Definition \ref{def:satisfaction}.

\begin{definition}
\label{def:wc-satisfaction}
For a weight constraint program $P$ and an interpretation
$I\subseteq\hb{P}$,
\begin{enumerate}
\item
a weight constraint $C$ of the form $l\leq\set{A=W_A,\naf B=W_B}\leq
u$ is satisfied in $I$ $\iff$
$l\leq\wsum{I}{A=W_A,\naf B=W_B}\leq u$,

\item
a weight constraint rule of the form $C_0\IF\rg{C_1}{,}{C_r}$ is
satisfied in $I$ $\iff$ $I\models C_0$ is implied by
$I\models C_1$, $\ldots$, and $I\models C_r$, and

\item
$I\models P$ $\iff$ $I\models r$ for every weight constraint rule $r\in P$.
\end{enumerate}
\end{definition}

The stable model semantics of normal programs \citeN{GL88:iclp} can be
generalized to the case of weight constraint programs using the
reduction devised for them by \citeN{SNS02:aij}.

\begin{definition}
\label{def:wc-reduct}
Given an interpretation $I$ for a weight constraint $C$ of the form
$l\leq\set{A=W_A,\naf B=W_B}\leq u$, the reduct $\GLred{C}{I}$ is the
constraint $l'\leq\{A=W_A\}$ where the lower bound
$l'=\max(0, l-\wsum{I}{\naf B=W_B})$.
\end{definition}

\begin{definition}
\label{def:wc-program-reduct}
For a weight constraint program $P$ and an interpretation
$I\subseteq\hb{P}$, the reduct $\GLred{P}{I}$ contains a reduced
weight constraint rule $h\IF \GLred{C_1}{I},\ldots, \GLred{C_r}{I}$
for each $C_0\IF C_1,\ldots,C_r\in P$ and $h\in A_0\isect I$
satisfying for all $i\in\eset{1}{r}$,
$\wsum{I}{A_i=W_{A_i},\naf B_i=W_{B_i}}\leq u_i$.
\end{definition}

It should be pointed out that $\GLred{P}{I}$ consists of {\em Horn
constraint rules} of the form $h\IF C_1,\ldots, C_r$, where $h$ is an
atom, each constraint $C_i$ contains only positive literals and has
only a lower bound condition. We say that a weight constraint program
$P$ is positive if all the rules in $P$ are Horn constraint
rules. Thus $\GLred{P}{I}$ is positive by definition.
The properties of minimal models carry over to the case of weight
constraint programs, too. Thus a {\em positive} weight constraint
program $P$ has a unique minimal model, the least model, $\lm{P}$, and
we can define stable models for weight constraints programs almost in
analogy to \system{smodels} programs.

\begin{definition}
\label{def:wc-sm}
An interpretation $M\subseteq\hb{P}$ for a weight constraint
program $P$ is a stable model of $P$ $\iff$
(i) $M\models P$ and (ii) $M=\lm{\GLred{P}{M}}$.
\end{definition}

This definition is only slightly different from Definition
\ref{def:sm} as $M\models P\iff M\models \GLred{P}{M}$ does not hold
generally for weight constraint programs --- making condition (i) in
Definition \ref{def:wc-sm} necessary.
However, if we consider the restricted language described in Section
\ref{section:syntax} and interpret the rules involved as weight
constraint rules (\ref{eq:wc-rule}), then Definitions \ref{def:sm} and
\ref{def:wc-sm} yield the same semantics as stated below. To this end,
we consider only choice rules (\ref{eq:choice-rule}) and weight rules
(\ref{eq:weight-rule}) without loss of generality.
\citeN{SNS02:aij} encode rules of these forms using the following
weight constraint rules:
\begin{multline}
\label{eq:choice-as-wcr}
\limit{0}{\rg{h_1=1}{,}{h_l=1}}\IF \\
\limit{n+m}{\rg{a_1=1}{,}{a_n=1},\rg{\naf b_1=1}{,}{\naf b_m=1}}
\end{multline}
\begin{equation}
\label{eq:weight-as-wcr}
\limit{1}{h=1}
\IF\limit{w}{\rg{a_1=w_{a_1}}{,}{a_n=w_{a_n}},
             \rg{\naf b_1=w_{b_1}}{,}{\naf b_m=w_{b_m}}}
\end{equation}

\begin{proposition}
\label{prop:same-semantics}
Let $P$ be an \system{smodels} program and $P_w$ its
representation as a weight constraint program.
Then for any interpretation $M\subseteq\hb{P}=\hb{P_w}$,
\begin{center}
$M=\lm{\GLred{P}{M}}$ $\iff$ $M\models P_w$ and $M=\lm{\GLred{P_w}{M}}$.
\end{center}
\end{proposition}

The proof of this proposition is given in \ref{appendix:proofs}.
\citeN{SNS02:aij} show how weight constraint programs can be
translated into {\em \system{smodels} programs} consisting only of
basic rules (\ref{eq:basic-rule}), choice rules (\ref{eq:choice-rule})
and weight rules (\ref{eq:weight-rule}).  The translation is highly
{\em modular} so that each weight constraint rule can be translated
independently of each other.
However, in order to keep the length of the translation linear, two
new atoms have to be introduced for each weight constraint appearing
in a program.

\begin{definition}
\label{def:tr-wc}
The translation $\trop{SNS}(C)$ of a weight constraint $C$ of the form
$l\leq\set{A=W_{A},\naf B=W_{B}}\leq u$ is translated into two weight
rules
\begin{eqnarray}
\label{eq:lower-bound}
\SAT{C}\IF\limit{l}{A=W_{A},\naf B=W_{B}}\END \\
\label{eq:upper-bound}
\UNSAT{C}\IF\limit{u+1}{A=W_{A},\naf B=W_{B}}\END
\end{eqnarray}
where $\SAT{C}$ and $\UNSAT{C}$ are new atoms specific to $C$.
\end{definition}

\begin{definition}
\label{def:tr-wcp-for-smodels}
Let $P$ be a weight constraint program and $f\not\in\hb{P}$ a new
atom. The translation of $P$ into an \system{smodels} program
$\trop{SNS}(P)$ consists of
\begin{enumerate}
\item
$\tr{SNS}{C}$ for each weight constraint $C$ appearing in $P$ and

\item
the following \system{smodels} rules introduced for each $C_0\IF
C_1,\ldots,C_r\in P$:
\begin{eqnarray}
\label{eq:make-choice}
\choice{A_0}\IF
  \SAT{C_1},\naf\UNSAT{C_1},\ldots,\SAT{C_r},\naf\UNSAT{C_r}\END \\
\label{eq:check-head1}
f\IF\naf\SAT{C_0},
        \SAT{C_1},\naf\UNSAT{C_1},\ldots,\SAT{C_r},\naf\UNSAT{C_r}\END \\ 
\label{eq:check-head2}
f\IF\UNSAT{C_0},
    \SAT{C_1},\naf\UNSAT{C_1},\ldots,\SAT{C_r},\naf\UNSAT{C_r}\END \\
\label{eq:check-head3}
\compute{\naf f}\END
\end{eqnarray}
\end{enumerate}
where $A_0$ is the set of positive default literals appearing in $C_0$.

The visible Herbrand base of $\trop{SNS}(P)$ is defined by
$\hbv{\trop{SNS}(P)}=\hbv{P}$.
\end{definition}

Let us then briefly explain intuitions underlying $\trop{SNS}$.
The rules given in (\ref{eq:lower-bound}) and (\ref{eq:upper-bound})
check whether the lower bound $l$ of the weight constraint $C$ is
satisfied the upper bound $u$ of $C$ is \emph{not} satisfied,
respectively, and then $\SAT{C}$ and $\UNSAT{C}$ can be inferred by
the rules accordingly.
The choice rule in (\ref{eq:make-choice}) makes
any subset of $A_0$ true if the body of the weight constraint rule is
satisfied in the sense of Definition \ref{def:wc-satisfaction}, i.e.,
all lower bounds and upper bounds are met. Finally, two basic rules in
(\ref{eq:check-head1}) and (\ref{eq:check-head2}) and the compute
statement in (\ref{eq:check-head3}) ensure the satisfaction of the
head constraint $C_0$ whenever the body $\rg{C_1}{,}{C_r}$ is
satisfied. A very tight correspondence of stable models is obtained
using the translation $\tr{SNS}{P}$.

\begin{theorem}
\label{theorem:tr-sns-faithful}
The translation function $\trop{SNS}$ is faithful, i.e.,
$P\lpeq{v}\tr{SNS}{P}$ holds for all weight constraint programs $P$.
\end{theorem}

The proof of the theorem can be found in \ref{appendix:proofs}.
Theorem \ref{theorem:tr-sns-faithful} and Definition
\ref{def:tr-wcp-for-smodels} imply together that the visible
equivalence of weight constraint programs can be reduced to that of
\system{smodels} programs using $\trop{SNS}$.

\begin{corollary}
\label{corollary:reduce-wcp-equivalence}
For all weight constraint programs $P$ and $Q$,
\begin{center}
$P\lpeq{v}Q \iff \trop{SNS}(P)\lpeq{v}\trop{SNS}(Q)$.
\end{center}
\end{corollary}

However, we have to address the degree of visibility of atoms in the
translation $\tr{SNS}{P}$ in order to apply the translation-based
method presented in Section \ref{section:translation}. Recalling the
limitations of the method, we should establish that $\tr{SNS}{P}$ and
$\tr{SNS}{Q}$ have enough visible atoms under some reasonable
assumptions about $P$ and $Q$. For the sake of simplicity, we will
only consider a relatively straightforward setting made precise in
Proposition \ref{prop:enough-visible-atoms}. Nevertheless, it implies
the applicability of our verification method to a substantial class of
weight constraint programs.

\begin{proposition}
\label{prop:enough-visible-atoms}
If $P$ is a weight constraint program such that $\hbh{P}=\emptyset$,
then $\tr{SNS}{P}$ has enough visible atoms.
\end{proposition}

\begin{proof}
Let $P$ be any weight constraint program such that
$\hbh{P}=\emptyset$, i.e., $\hbv{P}=\hb{P}$. Moreover, let us pick any
interpretation $\vis{I}\subseteq\hbv{P}$.  Since $\hbh{P}=\emptyset$
we have $\vis{I}=I$ so that $I$ is actually an interpretation for the
whole program.

Let us then consider any rule $C_0\IF\rg{C_1}{,}{C_r}\in P$ and its
translation under $\trop{SNS}$ as given in Definition
\ref{def:tr-wcp-for-smodels}.  Since $\hbv{\tr{SNS}{P}}=\hbv{P}$ by
definition and $I=\vis{I}$, the rules involved in the translation
contribute to $\eval{\hid{P}}{\vis{I}}$ as follows:
(\ref{eq:lower-bound})
is reduced to $\SAT{C_i}\IF\limit{l'_i}{}$ where $l'_i=\max(0,l_i-w_i)$ for
$w_i=\wsum{\vis{I}}{A_i=W_{A_i},\naf B_i=W_{B_i}}$;
(\ref{eq:upper-bound})
is reduced to $\UNSAT{C_i}\IF\limit{u'_i}{}$ where $u'_i=\max(0,(u_i+1)-w_i)$;
(\ref{eq:make-choice}) is dropped altogether as $\hid{(A_0)}=\emptyset$;
(\ref{eq:check-head1}) and (\ref{eq:check-head2}) remain intact
because they involve only hidden atoms; and (\ref{eq:check-head3}) is
dropped by definition.

Let us then verify that $\eval{\hid{\tr{SNS}{P}}}{\vis{I}}$ is a {\em
stratified} program. After inspecting the dependencies in the reduced
rules, we note that the hidden atoms in $\hb{\tr{SNS}{P}}$ can be
assigned to strata as follows: the atoms $\SAT{C}$ and $\UNSAT{C}$
associated with weight constraints $C$ belong to stratum 0 and $f$
belongs to stratum $1$.
Thus $\tr{SNS}{P}$ is essentially a stratified normal logic program
as the remainders of weight rules act as facts. Then
$\tr{SNS}{P}$ has a unique stable model \cite{ABW88}.
\end{proof}

By denying occurrences of hidden atoms in weight constraint programs,
we obtain a translation-based method for verifying weak equivalence.
Note that the requirement $\hbv{P}=\hbv{Q}$, i.e., $\hb{P}=\hb{Q}$ in
this case, can be easily met by extending the Herbrand bases of
programs as discussed in Section \ref{section:syntax}.

\begin{corollary}
\label{corollary:eq-testing}
Let $P$ and $Q$ two weight constraint programs such that
$\hbh{P}=\hbh{Q}=\emptyset$ and $\hbv{P}=\hbv{Q}$.
Then $P\lpeq{}Q$ $\iff$ $\trop{SNS}(P)\lpeq{v}\trop{SNS}(Q)$ $\iff$
the translations $\eqt(\tr{SNS}{P},\tr{SNS}{Q})$ and
$\eqt(\tr{SNS}{Q},\tr{SNS}{P})$ have no stable models.
\end{corollary}

It seems that hidden atoms can be tolerated in weight constraint
programs to some degree, but we skip such an extension of Corollary
\ref{corollary:eq-testing} for space reasons. Nevertheless, the result
established above enables us to implement the verification of weak
equivalence for the programs produced by \system{lparse}.


\section{Experiments}
\label{section:experiments}

The translation function $\eqt$ presented in Section
\ref{section:translation} has been implemented in C under the Linux
operating system. The translator which we have named \system{lpeq}
takes two logic programs $P$ and $Q$ as its input and produces the
translation $\eqt(P,Q)$ as its output.
The implementation assumes the internal file format of
\system{smodels} which enables us to use the front-end \system{lparse}
of \system{smodels} in conjunction with \system{lpeq}.%
\footnote{A textual human-readable output can also be produced on request.}
It is yet important to note that \system{lpeq} checks that the {\em
visible} Herbrand bases of the programs being compared are exactly the
same as insisted by $\lpeq{v}$. In practice, visible atoms are recognized
as those having a name in the symbol table of a program.
The latest version of \system{lpeq} is also prepared to deal with
programs involving {\em invisible} atoms, e.g., introduced by the
front-end \system{lparse} as discussed in Section
\ref{section:weightconstraint}. Before producing the translation
$\eqt(P,Q)$, the translator uses Tarjan's algorithm to find {\em
strongly connected components} for the dependency graph of $\hid{Q}$
when its checks that $\eval{\hid{Q}}{\vis{I}}$ is stratifiable for all
$\vis{I}\subseteq\hbv{Q}$.  An overapproximation is used in this
respect: all dependencies of invisible atoms are taken into account
regardless of $\vis{I}$.  A successful test guarantees that $Q$ has
enough visible atoms so that $\eqt(P,Q)$ works correctly.  Otherwise,
an error message is printed for the user.

The current implementation (\system{lpeq} version 1.17) is available%
\footnote{Please consult \url{http://www.tcs.hut.fi/Software/lpeq/}
for binaries and scripts involved.}
in the WWW. The files related with benchmark problems and experiments
reported in this section are also provided.
To assess the feasibility of \system{lpeq} in practice we performed
a number of tests with different test cases. The running times of the
\system{lpeq} approach were compared with those of a fictitious
approach, i.e., the \system{naive} one:
\begin{enumerate}
\item\label{item:compute-model}
Compute one stable model $M$ of $P$ not computed so far.

\item
Check whether $Q$ has a stable model $N$ such that $\vis{M}=\vis{N}$.
Stop if not.

\item
Continue from step \ref{item:compute-model}
until all stable models of $P$ have been enumerated.
\end{enumerate}
It is obvious that a similar check has to be carried out in the other
direction to establish $P\lpeq{v}Q$ in analogy to Corollaries
\ref{corollary:test-equivalence} and \ref{corollary:eq-testing}.

There is still room for optimization in both approaches.  If one finds
a counter-example in one direction, then $P\not\lpeq{v}Q$ is known to
hold and there is no need to do testing in the other direction except
if one wishes to perform a thorough analysis.  Since running times
seem to scale differently depending on the direction, we count always
running times in both directions.  However, one should notice that the
search for counter-examples in one direction is stopped immediately
after finding a counter-example.
Since the running times of \system{smodels} may also depend on the
order of rules in programs and literals in rules, we shuffle them
randomly.

\begin{figure}
{\small
\begin{enumerate}
\item[(a)] Place a queen on each column
\begin{verbatim}
negq(X,Y2) :- q(X,Y), d(X), d(Y), d(Y2), Y2 != Y.
q(X,Y) :- not negq(X,Y), not q(X,Y2): d(Y2): Y2 != Y , d(X), d(Y).
hide negq(X,Y).
\end{verbatim}
\item[(b)] Place a queen on each column using a choice rule
\begin{verbatim}
1 { q(X,Y):d(Y) } 1 :- d(X).
\end{verbatim}
\item[(c)] Place a queen on each row
\begin{verbatim}
negq(X2,Y) :- q(X,Y), d(X), d(Y), d(X2), X2 != X.
q(X,Y) :- not negq(X,Y), not q(X2,Y): d(X2): X2 != X , d(X), d(Y).
hide negq(X,Y).
\end{verbatim}
\item[(d)] Make sure that queens do not threaten each other
(same row or diagonal)
\begin{verbatim}
:- d(X), d(Y), d(X1), q(X,Y), q(X1,Y), X1 != X.
:- d(X), d(Y), d(X1), d(Y1), q(X,Y), q(X1,Y1), X != X1,  Y != Y1, 
   abs(X - X1) == abs(Y - Y1).
d(1..queens).
\end{verbatim}
\item[(e)] Make sure that queens do not threaten each other
(same column or diagonal)
\begin{verbatim}
:- d(X), d(Y), d(Y1), q(X,Y), q(X,Y1), Y1 != Y.
:- d(X), d(Y), d(X1), d(Y1), q(X,Y), q(X1,Y1), X != X1,  Y != Y1, 
   abs(X - X1) == abs(Y - Y1).
d(1..queens).
\end{verbatim}
\end{enumerate}
}
\caption{Encoding the $n$-\prob{queens} problem. \label{fig:queens}}
\end{figure}

In both approaches, the \system{smodels} system (version 2.28) is
responsible for the computation of stable models for programs that
are instantiated using the front-end \system{lparse} (version
1.0.13). In the \system{lpeq} approach, the total running time in one
direction is the running time needed by \system{smodels} for trying to
compute {\em one} stable model of the translation produced by
\system{lpeq}. The translation time is also taken into account
although it is negligible. 
The \system{naive} approach has been implemented as a shell script.  The
running time in one direction consists of the running time of
\system{smodels} for finding the necessary (but not necessarily all)
stable models of $P$ plus the running times of \system{smodels} for
testing that the stable models found are also stable models of
$Q$. These tests are realized in practice by adding
$\vis{M}\union\sel{\naf a}{a\in\hbv{Q}\setminus\vis{M}}$ as a compute
statement to $Q$.
It is worth noting that the \system{naive} approach does not test in any
way that the stable model $N$ of $Q$ with $\vis{M}=\vis{N}$ is unique. 
However, the set of benchmarks is selected in such a way that
programs have enough visible atoms and the correctness of the
\system{naive} approach is guaranteed.
All the tests reported in this section were run under the Linux 2.6.8
operating system on a 1.7GHz AMD Athlon XP 2000+ CPU with 1 GB of main
memory. As regards timings in test results, we report the sum of user
and system times.

\subsection{The $n$-Queens Benchmark}

\begin{table}[t]
\caption{Results for two equivalent logic programs ($n$-\prob{queens}).}
\label{queens1}
\begin{minipage}{\textwidth}
\begin{tabular}{rrrrrrrrr}
\hline\hline
$n$ & 
SM\footnote{Number of stable models for $Q_n^{x_1}$ and $Q_n^{x_2}$.} &
$t_{\mathrm{avg}}$\footnote{Average running time in seconds.} &
$t_{\mathrm{avg}}$ &
RAR\footnote{Ratio of average running times.} &
CP$_\mathrm{avg}$\footnote{Average number of choice points during the search.} &
CP$_\mathrm{avg}$ &
RI\footnote{Number of rules in the input: $|Q_n^{x_1}|+|Q_n^{x_2}|$.} &
RO\footnote{Number of rules in the output:
  $|\eqt(Q_n^{x_1},Q_n^{x_2})|+|\eqt(Q_n^{x_2},Q_n^{x_1})|$.} \\ 
&& \system{lpeq} & \system{naive} && \system{lpeq} & \system{naive} && \\
\hline
 1 &    1 & 0.000   &   0.080 &     -  &     0 &     0 & 7    & 28    \\
 2 &    0 & 0.000   &   0.051 &     -  &     0 &     0 & 38   & 130   \\
 3 &    0 & 0.003   &   0.051 & 17.000 &     0 &     0 & 124  & 384   \\
 4 &    2 & 0.019   &   0.120 & 6.316  &     0 &     2 & 300  & 884   \\
 5 &   10 & 0.042   &   0.454 & 10.810 &     5 &    18 & 600  & 1718  \\
 6 &    4 & 0.136   &   0.259 & 1.904  &    16 &    18 & 1058 & 2974  \\
 7 &   40 & 0.516   &   2.340 & 4.535  &    40 &    84 & 1708 & 4740  \\ 
 8 &   92 & 2.967   &   6.721 & 2.265  &   163 &   253 & 2584 & 7104  \\ 
 9 &  352 & 17.316  & 32.032  & 1.850  &   615 &   955 & 3720 & 10154 \\
10 &  724 & 99.866  & 90.694  & 0.908  &  2613 &  3127 & 5150 & 13978 \\ 
11 & 2680 & 617.579 & 451.302 & 0.731  & 11939 & 13662 & 6908 & 18664 \\ 
\hline\hline
\end{tabular}
\vspace{-2\baselineskip} 
\end{minipage}
\end{table}

Our first experiment was based on the $n$-\prob{queens} problem. We
verified the visible equivalence of three different formulations which
are variants of one proposed by \citeN[p. 260]{Niemela99:amai}. The
encoding $Q^{x_1}_n$ consists of parts (a) and (d) given in
Fig.~\ref{fig:queens} and is designed so that queens are placed
column-wise to the board.
The second program $Q^{x_2}_n$ consists of parts (b) and (d) given in
Fig.~\ref{fig:queens}, i.e., it is a variant of $Q^{x_1}_n$ where
the choice between placing or not placing a queen in a particular cell
of the chessboard is equivalently formulated using a choice rule
rather than plain basic rules. 
The third program $Q_n^y$, i.e., parts (c) and (e) given in
Fig.~\ref{fig:queens}, is an orthogonal version of $Q_n^{x_1}$ in which
queens are placed row-wise rather than column-wise. 

First we verified the visible equivalence of $Q_n^{x_1}$ and $Q_n^{x_2}$
and then that of $Q_n^{x_1}$ and $Q_n^{y}$ using both the
\system{lpeq} and the \system{naive} approaches. The number of queens
$n$ was varied from 1 to 11 and the verification task was repeated 100
times for each number of queens --- generating each time new randomly
shuffled versions of the programs involved. The results of these
experiments are shown in Tables \ref{queens1} and \ref{queens2},
respectively.
It appears that the \system{naive} approach becomes superior in the case
of two equivalent well-structured logic programs containing hidden
atoms (the atoms {\tt negq(X,Y)} are explicitly hidden) as programs
grow and the number of stable models increases.
Comparing the average running times from Tables \ref{queens1} and
\ref{queens2}, it can be seen that the difference in running times is
smaller in the case where the second program does not contain hidden
atoms. This can be seen as an indication that it is particularly the
translation of the hidden part $\trhid{\cdot}$ that increases the
running time of the \system{lpeq} approach. To investigate this
further, we verified the equivalence of $Q_n^{x_1}$ and $Q_n^{y}$
without declaring the atoms {\tt negq(X,Y)} hidden. The results
obtained from this experiment resembled our previous results in
\cite{JO02:jelia}, i.e., the \system{lpeq} approach performs somewhat
better than the \system{naive} one. Moreover, the average running
times of \system{naive} approach are approximately the same as with
hidden atoms, but the average running times for the \system{lpeq}
approach are significantly smaller.
The reason why the \system{naive} approach appears to be immune to
changes in the visibility of atoms is the following. In our encodings
of the $n$-\prob{queens} problem, the interpretation for hidden atoms
can be directly determined once the interpretation for visible part is
known. However, this is not necessarily the case in general and
finding the unique stable model for the hidden part can be more
laborious and time consuming as in our last benchmark to be described
in Section \ref{section:knapsack}.

\begin{table}[t]
\caption{Results for two equivalent logic programs ($n$-\prob{queens}).}
\label{queens2}
\begin{minipage}{\textwidth}
\begin{tabular}{rrrrrrrrr}
\hline\hline
$n$ & 
SM\footnote{Number of stable models for $Q_n^{x_1}$ and $Q_n^{y}$.} &
$t_{\mathrm{avg}}$\footnote{Average running time in seconds.} &
$t_{\mathrm{avg}}$ &
RAR\footnote{Ratio of average running times.} &
CP$_\mathrm{avg}$\footnote{Average number of choice points during the search.} &
CP$_\mathrm{avg}$ &
RI\footnote{Number of rules in the input: $|Q_n^{x_1}|+|Q_n^{y}|$.} &
RO\footnote{Number of rules in the output:
  $|\eqt(Q_n^{x_1},Q_n^{y})|+|\eqt(Q_n^{y},Q_n^{x_1})|$.} \\ 
&& \system{lpeq} & \system{naive} && \system{lpeq} & \system{naive} && \\
\hline
 1  &    1 &    0.000 &   0.080 &    - &     0 &     0 & 4    & 30    \\
 2  &    0 &    0.000 &   0.050 &    - &     0 &     0 & 36   & 146   \\
 3  &    0 &    0.007 &   0.052 & 7.43 &     0 &     0 & 136  & 478   \\
 4  &    2 &    0.020 &   0.124 & 6.20 &     0 &     2 & 344  & 1146  \\
 5  &   10 &    0.052 &   0.473 & 9.09 &     4 &    18 & 700  & 2270  \\
 6  &    4 &    0.169 &   0.281 & 1.66 &    16 &    18 & 1244 & 3970  \\
 7  &   40 &    0.815 &   2.583 & 3.17 &    38 &    84 & 2016 & 6366  \\
 8  &   92 &    5.994 &   7.531 & 1.26 &   176 &   263 & 3116 & 9578  \\
 9  &  352 &   35.900 &  36.836 & 1.03 &   603 &   955 & 4404 & 13726 \\
 10 &  724 &  238.726 & 110.109 & 0.46 &  2734 &  3243 & 6100 & 18930 \\
 11 & 2680 & 1521.730 & 565.029 & 0.37 & 12210 & 13927 & 8184 & 25310 \\
\hline\hline
\end{tabular}
\vspace{-2\baselineskip} 
\end{minipage}
\end{table}

We chose the pairs of programs $(Q_n^{x_1},Q_n^{x_2})$ and
$(Q_n^{x_1},Q_n^{y})$ for our experiments in order to to see if the
two approaches would perform differently depending on whether a {\em
local change} (a choice rule is used instead of basic rules) or a {\em
global change} (an orthogonal encoding is introduced) is made in the
encoding. However, our test results show no clear indication in either
direction.
Furthermore, we decided to test non-equivalent pairs of $n$-\prob{queens}
programs. To this end, we dropped $n$ random rules from $Q_n^{y}$, and
verified the equivalence of $Q_n^{x_1}$ and the modified version of
$Q_n^{y}$ by selecting only non-equivalent pairs (both with and
without hidden atoms). The results turned out to be very similar to
the results that were obtained for equivalent program pairs.
In all our $n$-\prob{queens} experiments the number of {\em choice
points} (i.e., the number of choices made by \system{smodels} while
searching for stable models for the translation) is slightly smaller
in the \system{lpeq} approach than in the \system{naive} one.
Thus it seems that verifying the equivalence of logic programs using
\system{lpeq} leads to smaller search space, but the eventual
efficiency can vary as far as time is concerned.

\subsection{Random $3$-SAT and Graph Problems}


\begin{figure}[t]
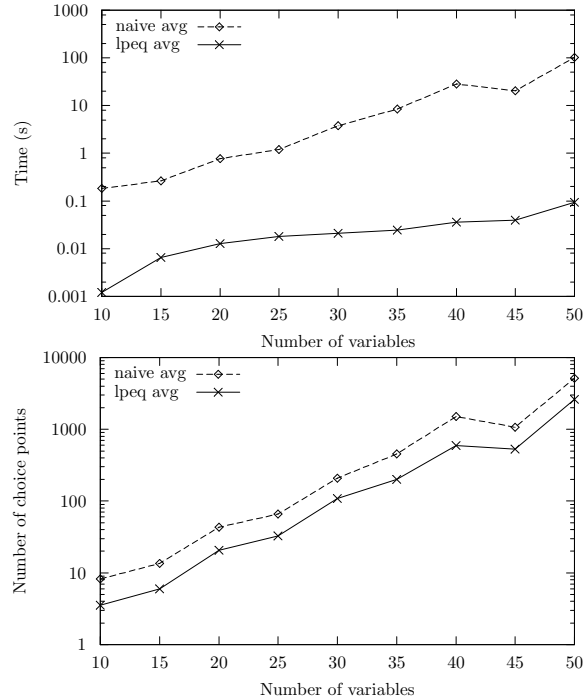

\begin{center}
\epsfig{file=fig1.ps,width=0.6\textwidth} \\
\epsfig{file=fig2.ps,width=0.6\textwidth}
\end{center}
\caption{Average running times and numbers of choice points for
  random 3-\prob{sat} instances with the ratio $c/v=4$. 
         \label{const_ratio}}
\end{figure}

We also performed some tests with randomly generated logic programs.
We generated logic programs that solve an instance of a random
3-\prob{sat} problem with a constant clauses to variables ratio
$c/v=4$. Such instances are typically satisfiable, but so close to the
{\em phase transition point} (approximately 4.3) that finding models
is already demanding for SAT solvers. To simulate a sloppy programmer
making mistakes, we dropped one random rule from each program. Due to
non-existence of hidden atoms, we checked the weak equivalence of the
modified program and the original program to see if making such a
mistake affects stable models or not.
As a consequence, the pairs of programs included both equivalent and
nonequivalent cases.
When $c/v=4$, approximately 45--60\% of the program pairs were
equivalent. This does not seem to depend much on the problem size
(measured in the number of variables in the problem) within problem
sizes used in the experiments. With smaller values of $c/v$ the
percentage of equivalent program pairs is lower but for larger values
of $c/v$ the percentage grows up to 90\%.  
In the first experiment with random 3-\prob{sat} programs, we varied
the number of variables $v$ from 10 to 50 with steps of 5. For each
number of variables we repeated the test 100 times and generated each
time a new random instance. The average running times and the average
number of choice points for both approaches are shown in Figure
\ref{const_ratio}. These results indicate that the \system{lpeq}
approach is significantly faster than the \system{naive} one. The
difference increases as program instances grow. The number of choice
points is also lower in the former approach on an average.


\begin{figure}[t]
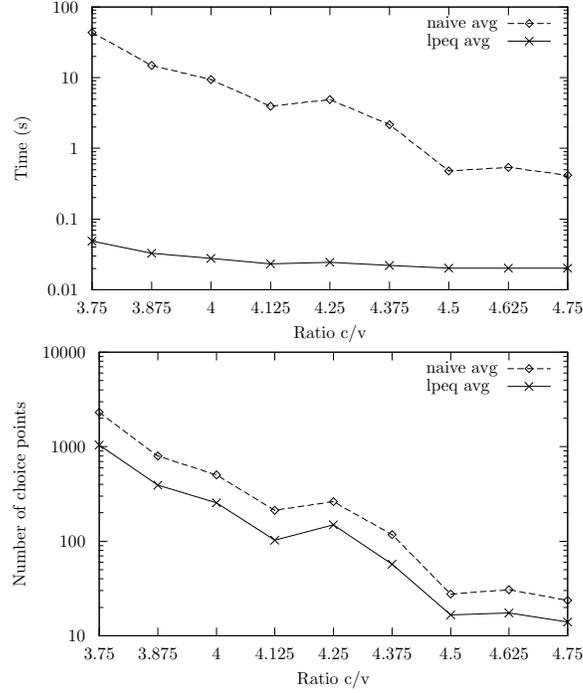

\begin{center}
\epsfig{file=fig3.ps,width=0.6\textwidth} \\
\epsfig{file=fig4.ps,width=0.6\textwidth}
\end{center}
\caption{Average running times and numbers of choice points for
random 3-\prob{sat} instances with fixed $v=40$ and varying ratio
$c/v$.
         \label{varying_ratio}}
\end{figure}

In the second experiment with random 3-\prob{sat} instances we
generated programs as in the previous experiment, but we kept the
number of variables constant, $v=40$, and varied the ratio $c/v$ from
3.75 to 4.75 with steps of 0.125. For each value of the ratio $c/v$,
we repeated the test 100 times generating each time a new random
instance. The motivation behind this experiment was to see how the
\system{lpeq} approach performs compared to the \system{naive} one as
the programs change from almost always satisfiable (many stable
models) to almost always unsatisfiable (no stable models).
The averages of running times and numbers of choice points are presented
in Figure \ref{varying_ratio} for both approaches. For low values of the
ratio $c/v$, the \system{lpeq} approach is significantly better than the
\system{naive} one like previously. As the ratio increases, the
performance of the \system{naive} approach gradually improves, but
\system{lpeq} is still better. The average number
of choice points is also lower in the \system{lpeq} approach.

\begin{figure}[t]
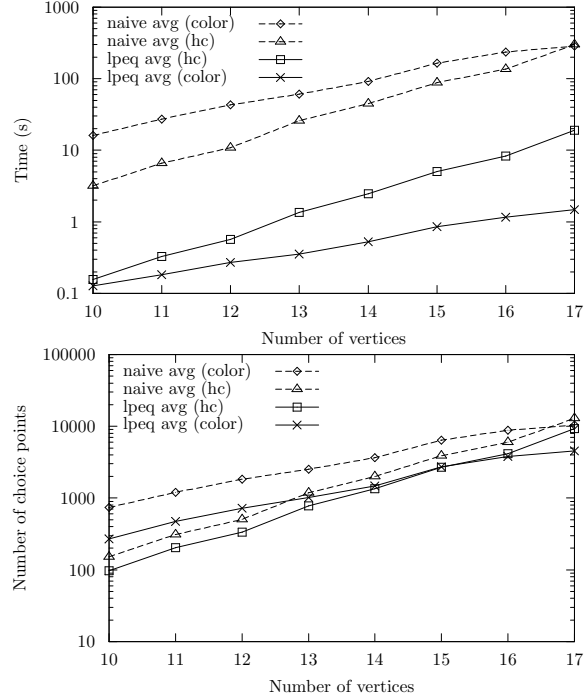

\begin{center}
\epsfig{file=fig5.ps,width=0.6\textwidth} \\
\epsfig{file=fig6.ps,width=0.6\textwidth}
\caption{Average running times and average number of choice points  for
  4-\prob{coloring} random planar graphs and 
         finding \prob{Hamiltonian circuits}.
         \label{graph_hc_color}}
\end{center}
\end{figure}


We also combined structured logic programs with randomness.  We used
two graph problems formalized with rules by
\citeN[p. 262]{Niemela99:amai}: the problems of $n$-\prob{coloring} of
a graph with $n$ colors and finding a \prob{Hamiltonian circuit} for a
graph. Using the Stanford GraphBase library, we generated random planar
graphs with $v$ vertices where $v$ ranges from 10 to 17 and
instantiated the respective logic programs for $4$-\prob{coloring} and
\prob{Hamiltonian circuit} by invoking \system{lparse}.
As in the preceding experiments with random 3-\prob{sat} programs, the
second program for equivalence testing was obtained by dropping one
random rule from the one instantiated by \system{lparse}.  The tests
were repeated 100 times for each value of $v$ using a new random planar
graph every time.
The average running times and the average number of choice points for
both experiments are presented in Figure \ref{graph_hc_color}. In both
experiments the \system{lpeq} approach is significantly better than
the \system{naive} approach, though running times differ more in the
4-\prob{coloring} problem.  The numbers of choice points
vary as before.

\subsection{Knapsack}
\label{section:knapsack}

\begin{figure}[b]
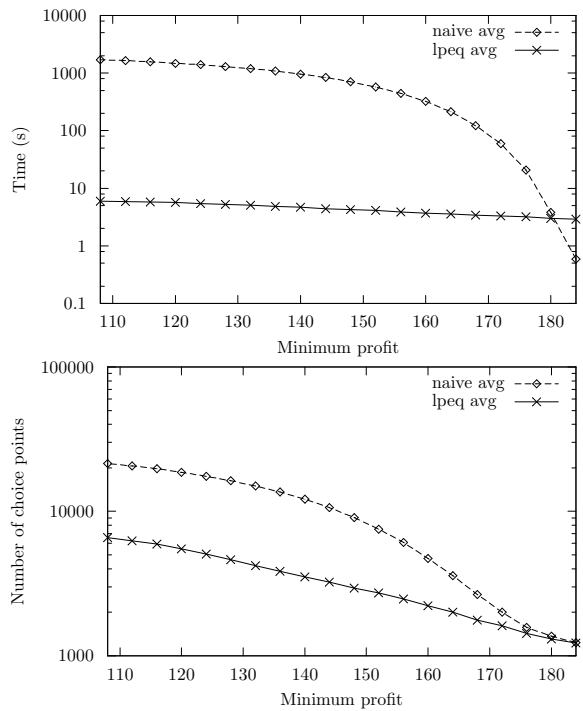

\begin{center}
\epsfig{file=fig7.ps,width=0.6\textwidth} \\
\epsfig{file=fig8.ps,width=0.6\textwidth}
\caption{Average running times and average number of choice points for
\prob{knapsack}.
         \label{graph_knapsack}}
\end{center}
\end{figure}

Finally, we used the \prob{knapsack} problem whose encoding involves
weight constraints. Here the objective was to test the performance of
the translation-based approach when programs involve a substantial number
of hidden atoms and the verification of equivalence requires the
property of having enough visible atoms as stated in Corollary
\ref{corollary:eq-testing}. It should be stressed that the previous
version of \system{lpeq} (1.13) and the corresponding translation
presented in \cite{JO02:jelia} do not cover such programs.
In the knapsack problem, there are $n$ types of items, each item of
type $i$ has size $w_i$ and profit $c_i$. The goal is to fill the
knapsack with $X_i$ items of type $i$ so that the maximum size $W$ is
not exceeded and the minimum profit $C$ is gained, i.e.,
$$\sum^{n}_{i=1}X_i \cdot w_i\leq W\;
\mathrm{and}\;\;\sum^{n}_{i=1}X_i \cdot c_i \geq C.$$ 
We decided to use an encoding of the \prob{knapsack} problem
proposed by \citeN{DFP05:cilc} using the same weights and costs.
An instance of the encoding is denoted by $\mathrm{KS}(W,C)$ where the
parameters $W$ and $C$ are as above.
We considered the visible equivalence of programs $\mathrm{KS}(127,C)$
and $\mathrm{KS}(127,C-1)$ for the values of $C$ in the sequence
$184, 180, \ldots, 104, 100$. 
The starting value $C=184$ was selected, since it is the highest
possible value for $\mathrm{KS}(127,C)$ to have stable models. As $C$
decreases, the number of stable models possessed by
$\mathrm{KS}(127,C)$ grows.
For each value of $C$ we generated 10 randomly shuffled versions of
$\mathrm{KS}(127,C)$ and  $\mathrm{KS}(127,C-1)$. 
The programs $\mathrm{KS}(127,C)$ and $\mathrm{KS}(127,C-1)$ are
always visibly non-equivalent as stable models of $\mathrm{KS}(127,C)$
are also stable models of $\mathrm{KS}(127,C-1)$ up to visible parts,
but not vice versa, because of weights used in \cite{DFP05:cilc}.

The averages of running times and numbers of choice points
for the \prob{knapsack} problem are presented in Figure
\ref{graph_knapsack}. 
It is worth noting that the total running time is dominated by the 
direction that does not yield a counter-example.
However, the \system{lpeq} approach is also significantly faster than
the \system{naive} one in the direction that actually yields
counter-examples.


\section{Conclusion}
\label{section:conclusion}

In this article, we propose a translation-based approach for verifying
the equivalence of logic programs under the stable model semantics.
The current translation $\eqt(P,Q)$ and its implementation
\system{lpeq} cover the types of rules supported by the
\system{smodels} search engine which provide the basic knowledge
representation primitives.
More general forms of rules implemented in the front-end
\system{lparse} are also covered by \system{lpeq}. This is partially
achieved by \system{lparse} itself as it expresses high-level
constructs using the primitives of the engine. However, the task of
verifying equivalence is complicated considerably since
\system{lparse} may have to introduce hidden atoms.  To this end, the
newest version of \system{lpeq} includes a proper support for hidden
atoms so that it can be used to verify {\em visible equivalence} of
\system{smodels} programs (denoted $\lpeq{v}$) rather than ordinary
weak equivalence (denoted $\lpeq{}$). The underlying theory around the
property of having enough visible atoms is developed in Section
\ref{section:equivalence} and we consider these ideas as a significant
extension to the original translation-based approach presented in
\cite{JO02:jelia}.

Our conclusion of the experiments reported in Section
\ref{section:experiments} is that the translation-based approach can
really be useful in practice. In many cases, the number of choice
points and time needed for computations is less than in the
\system{naive} cross-checking approach.
To the best of our understanding, this is because the translation
$\eqt(P,Q)$ provides an explicit specification of a counter-example
that guides the search performed by \system{smodels}. Such
coordination is not possible in the \system{naive} approach where the
stable models of $P$ and $Q$ are computed separately and
cross-checked.
However, if the programs being compared are likely to have few stable
models or no stable models at all, we expect that the \system{naive}
approach becomes superior to ours. Recall that $P$ is included in the
translation $\eqt(P,Q)$ which has no stable models in the case that
$P$ has no stable models. The \system{naive} approach may also be
better off when programs turn out to be equivalent and the
verification task boils down to establishing the correspondence of
stable models.

As regards future work, there are several issues to be addressed.
\begin{itemize}
\item
The current translation and its implementation \system{lpeq} do not
cover minimize/maximize statements that are nevertheless supported by
the \system{smodels} search engine. Basically, one can deal with
optimization on two levels.
The first is to verify the equivalence of programs without
optimization statements which should intuitively imply equivalence in
the presence of the same optimization statements expressed in terms of
visible atoms. The second approach is the fully general one that
allows differences in the non-optimal models of the programs being
compared and in the formulation of optimization statements as there
may be several formulations that are effectively equivalent.

\item
Other notions of equivalence --- such as the stronger notion of
equivalence proposed by \citeN{LPV01:acmtocl} --- should be covered by
devising and implementing suitable translations.  Some translations in
this respect have already been presented by \citeN{Turner03:tplp} and
\citeN{EFTW04:lpnmr}. However, the visibility aspects of these
relations have not been fully analyzed so far.

\item
The current implementation provides already a reasonably good support
for invisible atoms, since those introduced by \system{lparse} can be
dealt with. However, the notion of stratification used by
\system{lpeq} is very cautious and we should also pursue other natural
classes of programs that have enough visible atoms. One obvious
question in this respect is whether the property of having enough
visible atoms is {\em preserved} by \system{lparse}.

\item
The case of disjunctive logic programs is also interesting, as
efficient implementations are available: \system{dlv}
\cite{LPFEGPS06:acmtocl} and \system{gnt} \cite{JNSSY06:acmtocl}. The
latter uses \system{smodels} for actual computations in analogy to the
translation-based approach followed by this paper. In
\cite{OJ04:lpnmr} we extend the translation-based approach to the
disjunctive case. The respective implementation for disjunctive
programs, namely \system{dlpeq}, is reported in \cite{JO04:lpnmr}.
For now, invisible atoms are not supported by \system{dlpeq} and it is
interesting to see whether the concept of having enough visible atoms
lifts to the disjunctive case in a natural way.
\end{itemize}


\section*{Acknowledgments}

The authors wish to thank Michael Gelfond, Stefan Woltran, and
anonymous referees for their comments and suggestions for
improvement. This research has been partially supported by the Academy
of Finland (under Projects \#53695 {\em ``Applications of Rule-Based
Constraint Programming''}, and \#211025 {\em ``Applications of
Constraint Programming Techniques''}) and the European Commission
(under contract IST-FET-2001-37004 {\em ``Working Group on Answer Set
Programming''}). The second author gratefully acknowledges financial
support from Helsinki Graduate School in Computer Science and
Engineering, Nokia Foundation, and Finnish Cultural Foundation.

\appendix


\section{Proofs}
\label{appendix:proofs}

\begin{proof}[Proof of Lemma \ref{lemma:uniqueness}]
Suppose that $M\in\sm{P}$, i.e.\ $M=\lm{\GLred{P}{M}}$ and
$M\models\compst{P}$. To prove
$\hid{M}\in\sm{\eval{\hid{P}}{\vis{M}}}$, let us establish first
that $\hid{M}\models\GLred{(\eval{\hid{P}}{\vis{M}})}{\hid{M}}$.
Assuming the contrary, some rule
$r\in\GLred{(\eval{\hid{P}}{\vis{M}})}{\hid{M}}$
must be falsified by $\hid{M}$. Since basic rules and constraint rules
are special cases of weight rules (c.f.\ discussion after Definition
\ref{def:rules}), it is sufficient to consider only rules $r$ of two
types: weight rules and choice rules.
\begin{itemize}
\item
If a weight rule $h\IF\limit{w_1}{\hid{A}=W_{\hid{A}}}$ in
$\GLred{(\eval{\hid{P}}{\vis{M}})}{\hid{M}}$ is falsified by $\hid{M}$,
we have $\hid{M}\not\models h$ and
$w_1\leq\wsum{\hid{M}}{\hid{A}=W_{\hid{A}}}$
in which
$w_1=\max(0,w_2-\wsum{\hid{M}}{\naf\hid{B}=W_{\hid{B}}})$
is related with a rule
$h\IF\limit{w_2}{\hid{A}=W_{\hid{A}},\naf\hid{B}=W_{\hid{B}}}$
included in $\eval{\hid{P}}{\vis{M}}$. Thus
$w_2\leq
 \wsum{\hid{M}}{\hid{A}=W_{\hid{A}},\naf\hid{B}=W_{\hid{B}}}$.
Then the definition of $\eval{\hid{P}}{\vis{M}}$ and that of $w_1$
in terms of the bound $w_2$ imply that
$w_2=\max(0,w-\wsum{\vis{M}}{\vis{A}=W_{\vis{A}},\naf\vis{B}=W_{\vis{B}}})$
for some weight rule $h\IF\limit{w}{A=W_A,\naf B=W_B}$ of $P$.
By combining weight sums on the basis of $M=\hid{M}\union\vis{M}$, we
obtain $w\leq\wsum{M}{A=W_A,\naf B=W_B}$.
On the other hand, the reduct $\GLred{P}{M}$ contains a weight rule
$h\IF\limit{w_3}{A=W_A}$ where $w_3=\max(0,w-\wsum{M}{\naf B=W_B})$.
It follows that $w_3\leq\wsum{M}{A=W_A}$ and $M\not\models h$.
A contradiction, since $M\models\GLred{P}{M}$ holds for $M$.

\item
A choice rule $\choice{\hid{H}}\IF\hid{A},\naf\hid{B}$ cannot
be falsified by definition, a contradiction.
\end{itemize}
Hence $\hid{M}\models\GLred{(\eval{\hid{P}}{\vis{M}})}{\hid{M}}$ and it
remains to establish the minimality of $\hid{M}$ with respect to this
property. Suppose there is
$M'\models\GLred{(\eval{\hid{P}}{\vis{M}})}{\hid{M}}$
such that $M'\subset\hid{M}$.  Using $M'$ we define an interpretation
$N=\vis{M}\union M'$ so that $\vis{N}=\vis{M}$,
$\hid{N}=M'\subset\hid{M}$, and
$\hid{N}\models\GLred{(\eval{\hid{P}}{\vis{M}})}{\hid{M}}$ by
definition.
Let us then assume that $N\not\models\GLred{P}{M}$, i.e., there is
some rule $r$ of the reduct $\GLred{P}{M}$ not satisfied by $N\subset
M$.  As above, it is sufficient to consider the contribution of weight
rules and choice rules to $\GLred{P}{M}$.
\begin{itemize}
\item
If $r$ is a weight rule $h\IF\limit{w_2}{A=W_A}$ in $\GLred{P}{M}$,
then $N\not\models h$ and $w_2\leq\wsum{N}{A=W_A}$ holds for
$w_2=\max(0,w-\wsum{M}{\naf B=W_B})$ and some weight rule
$h\IF\limit{w}{A=W_A,\naf B=W_B}$ of $P$.  Since $\vis{M}$ and
$\vis{N}$ coincide, we obtain
\begin{equation}
\label{eq:weight-dependency}
\begin{array}{rcl}
w & \leq & \wsum{N}{A=W_A}+\wsum{M}{\naf B=W_B} \\
  & =    & \wsum{\hid{N}}{\hid{A}=W_{\hid{A}}}+
           \wsum{\hid{M}}{\naf\hid{B}=W_{\hid{B}}}+ \\
  &      &
  \wsum{\vis{M}}{\vis{A}=W_{\vis{A}},\naf\vis{B}=W_{\vis{B}}}.
\end{array}
\end{equation}
Two cases arise.
(i)
If $h\in\hbh{P}$, then $\eval{\hid{P}}{\vis{M}}$ contains a rule
$h\IF\limit{w_3}{\hid{A}=W_{\hid{A}},\naf\hid{B}=W_{\hid{B}}}$ where
$w_3=
 \max(0,w-\wsum{\vis{M}}{\vis{A}=W_{\vis{A}},\naf\vis{B}=W_{\vis{B}}})$.
It follows by (\ref{eq:weight-dependency}) that
$w_3\leq
 \wsum{\hid{N}}{\hid{A}=W_{\hid{A}}}+
 \wsum{\hid{M}}{\naf\hid{B}=W_{\hid{B}}}$.
Moreover, the reduct $\GLred{(\eval{\hid{P}}{\vis{M}})}{\hid{M}}$
includes a rule $h\IF\limit{w_4}{\hid{A}=W_{\hid{A}}}$ where
$w_4=\max(0,w_3-\wsum{\hid{M}}{\naf\hid{B}=W_{\hid{B}}})$.
Thus $w_4\leq\wsum{\hid{N}}{\hid{A}=W_{\hid{A}}}$ holds so that
$N\not\models h$ and $h\in\hbh{P}$ imply $\hid{N}\not\models r$,
a contradiction with
$\hid{N}\models\GLred{(\eval{\hid{P}}{\vis{M}})}{\hid{M}}$.
(ii)
If $h\in\hbv{P}$, then the definition of $N$ implies $M\not\models
h$. Moreover $N\subseteq M$ implies
$\wsum{N}{A=W_A}\leq\wsum{M}{A=W_A}$ so that
$w_2\leq\wsum{M}{A=W_A}$. Thus $M\not\models r$ and
$M\not\models\GLred{P}{M}$ which contradicts the fact that
$M=\lm{\GLred{P}{M}}$.

\item
If $r$ is a basic rule $h\IF A$ associated with a choice rule
$\choice{H}\IF A,\naf B$ of $P$, then $h\in H$, $M\models h$,
$M\models\naf B$, $N\not\models h$, and $N\models A$.
Now $h\in\hbv{P}$ is impossible as $\vis{M}=\vis{N}$,
$M\models h$, and $N\not\models h$. Hence $h\in\hbh{P}$
is necessarily the case and $\hid{H}\neq\emptyset$.
Moreover, $\vis{M}=\vis{N}$, $N\models A$, and $M\models\naf B$ imply
that $\vis{M}\models\vis{A}\union\naf\vis{B}$. Thus
$\choice{\hid{H}}\IF\hid{A},\naf\hid{B}$ is included in
$\eval{\hid{P}}{\vis{M}}$.
In addition, $M\models\naf B$ and $M\models h$ imply
$\hid{M}\models\naf\hid{B}$ and $\hid{M}\models h$ so that
$h\IF\hid{A}$ is included in
$\GLred{(\eval{\hid{P}}{\vis{M}})}{\hid{M}}$.
Finally, we obtain $\hid{N}\models\hid{A}$, $\hid{N}\not\models h$ and
$\hid{N}\not\models r$ from $N\models A$ and $N\not\models
h$. A contradiction.
\end{itemize}
To conclude the analysis above, it must be the case that
$N\models\GLred{P}{M}$. Since $N\subset M$, this contradicts the
fact that $M$ is a minimal model of $\GLred{P}{M}$. Thus $\hid{M}$ is
necessarily a minimal model of
$\GLred{(\eval{\hid{P}}{\vis{M}})}{\hid{M}}$,
i.e., a stable model of $\eval{\hid{P}}{\vis{M}}$.
\end{proof}


\begin{proof}[Proof of Proposition \ref{prop:least-model-of-reduct}]
We prove the given four claims depending on conditions (i)
$M=\lm{\GLred{P}{M}}$, (ii)
$\hid{N}=\lm{\GLred{(\eval{\hid{Q}}{\vis{M}})}{\hid{N}}}$, and (iii)
$L=\lm{\GLred{Q}{N}}$.
Let us define $J=\lm{\GLred{\eqt(P,Q)}{I}}$ for more concise notation.
It is clear that $J\models\GLred{\eqt(P,Q)}{I}$ holds.
\vspace{1\baselineskip}

\noindent
\textbf{Claim \ref{item:least-model-of-reduct1}:}
$J\isect\hb{P}=\lm{\GLred{P}{M}}$.
\vspace{1\baselineskip}

$(\supseteq)$
Since $\GLred{P}{M}\subseteq\GLred{\eqt(P,Q)}{I}$ by Lemma
\ref{lemma:reduct}, also $J\models\GLred{P}{M}$ holds. Then
$J\isect\hb{P}\models\GLred{P}{M}$ as $\GLred{P}{M}$ is based on
$\hb{P}$. Thus $\lm{\GLred{P}{M}}$ is contained in $J\isect\hb{P}$.

$(\subseteq)$
Now $\lm{\GLred{P}{M}}\models\GLred{P}{M}$ holds. Then define
an interpretation
$K=\lm{\GLred{P}{M}}\union\renh{\hbh{Q}}\union\ren{\hb{Q}}\union\set{c,d,e}$
for which $K\models\GLred{\eqt(P,Q)}{I}$ holds trivially by Lemma
\ref{lemma:reduct}. Thus $J\subseteq K$ and
$K\isect\hb{P}=\lm{\GLred{P}{M}}$ imply
$J\isect\hb{P}\subseteq\lm{\GLred{P}{M}}$.
\vspace{1\baselineskip}

\noindent
\textbf{Claim \ref{item:least-model-of-reduct2}:}
If (i), then
$J\isect\renh{\hbh{Q}}=\renh{\lm{\GLred{(\eval{\hid{Q}}{\vis{M}})}{\hid{N}}}}$.
\vspace{1\baselineskip}

Assuming (i) we obtain
$M=I\isect\hb{P}=\lm{\GLred{P}{M}}=J\isect\hb{P}$ by Claim
\ref{item:least-model-of-reduct1}.

($\supseteq$)
Let us assume that
$J\not\models\renh{(\GLred{(\eval{\hid{Q}}{\vis{M}})}{\hid{N}})}$.  In
this respect, it is sufficient to consider only cases where weight rules
and choice rules belong to the reduct.
\begin{itemize}
\item
Suppose there is a weight rule
$h\IF\limit{w_1}{\hid{A}=W_{\hid{A}},\naf\hid{B}=W_{\hid{B}}}\in
 \eval{\hid{Q}}{\vis{M}}$
where $h\in\hbh{Q}$ and
$w_1=\max(0,w-\wsum{\vis{M}}{\vis{A}=W_{\vis{A}},\naf\vis{B}=W_{\vis{B}}})$
is obtained from $h\IF\limit{w}{A=W_A,\naf B=W_B}\in Q$.
Then the rule $\renh{h}\IF\limit{w_2}{\renh{\hid{A}}=W_{\renh{\hid{A}}}}$
is in $\renh{(\GLred{(\eval{\hid{Q}}{\vis{M}})}{\hid{N}})}$ and
$w_2=\max(0,w_1-\wsum{\hid{N}}{\naf\hid{B}=W_{\hid{B}}})$.  Since
this rule is falsified under $J$, we have $J\not\models\renh{h}$ and
$w_2\leq\wsum{J}{\renh{\hid{A}}=W_{\renh{\hid{A}}}}$. Using the
definitions of $w_2$ and $w_1$, we obtain an inequality
\begin{equation}
\label{eq:wsums}
\begin{array}{rcl}
w & \leq &
\wsum{J}{\renh{\hid{A}}=W_{\renh{\hid{A}}}}+
\wsum{\hid{N}}{\naf\hid{B}=W_{\hid{B}}}+ \\
&&
\wsum{\vis{M}}
     {\vis{A}=W_{\vis{A}},\naf\vis{B}=W_{\vis{B}}}.
\end{array}
\end{equation}
On the other hand, there is a rule
$r=\renh{h}\IF\limit{w_3}{\renh{\hid{A}}=W_{\renh{\hid{A}}},
                          \vis{A}=W_{\vis{A}}}$
with $w_3=\max(0,w-\wsum{N}{\naf B=W_B})$
in $\GLred{\eqt(P,Q)}{I}$ by Lemma \ref{lemma:reduct}. Since
$N=\vis{M}\union\hid{N}$ by definition,
we obtain
$w_3\leq\wsum{J}{\renh{\hid{A}}=W_{\renh{\hid{A}}}}+
        \wsum{\vis{M}}{\vis{A}=W_{\vis{A}}}$
from the definition of $w_3$ and (\ref{eq:wsums}).
As $M=J\isect\hb{P}$, we know that $\vis{M}=J\isect\hbv{Q}$ and
$w_3\leq
 \wsum{J}{\renh{\hid{A}}=W_{\renh{\hid{A}}},\vis{A}=W_{\vis{A}}}$.
Thus $J\not\models r$ and $J\not\models\GLred{\eqt(P,Q)}{I}$
which contradicts the choice of $J$ in the beginning of this proof.

\item
Suppose there is a choice rule
$\choice{\hid{H}}\IF\hid{A},\naf\hid{B}\in \eval{\hid{Q}}{\vis{M}}$ so
that $\hid{H}\neq\emptyset$ and
$\vis{M}\models\vis{A}\union\naf\vis{B}$ hold for a rule
$\choice{H}\IF A,\naf B\in Q$. Consider any $h\in\hid{H}$.
If $\hid{N}\models h$, $\hid{N}\models\naf\hid{B}$, and
$\vis{M}\models\naf\vis{B}$, there is a rule
$\renh{h}\IF\renh{\hid{A}}$ included in
$\renh{(\GLred{(\eval{\hid{Q}}{\vis{M}})}{\hid{N}})}$.  Assuming that
this rule is falsified by $J$ implies that $J\not\models\renh{h}$ and
$J\models\renh{\hid{A}}$.
Since $N=\vis{M}\union\hid{N}$ by definition, we have $N\models\naf
B$.  Together with $\hid{N}\models h$, this implies that there is a
rule $r=\renh{h}\IF\renh{\hid{A}},\vis{A}$ in $\GLred{\eqt(P,Q)}{I}$
by Lemma \ref{lemma:reduct}. Since $\vis{M}=J\isect\hbv{Q}$ as above,
we obtain $J\models\vis{A}$ so that $J\not\models r$ and
$J\not\models\GLred{\eqt(P,Q)}{I}$.  A contradiction regardless of the
choice of $h$.
\end{itemize}

Thus $J\models\renh{(\GLred{(\eval{\hid{Q}}{\vis{M}})}{\hid{N}})}$
follows and $\renh{\lm{\GLred{(\eval{\hid{Q}}{\vis{M}})}{\hid{N}}}}$ is
necessarily contained in $J\isect\renh{\hb{Q}}$.

($\subseteq$)
Define an interpretation
$K=M\union
   \renh{\lm{\GLred{(\eval{\hid{Q}}{\vis{M}})}{\hid{N}}}}\union
   \ren{\hb{Q}}\union
   \set{c,d,e}$.
Since (i) is assumed, it is clear that $K\models\GLred{P}{M}$ but the
satisfaction of rules addressed in Items
\ref{item:reduced-Q-h-begin}--\ref{item:reduced-Q-h-end}
of Lemma \ref{lemma:reduct} must be
verified. A case analysis follows.
\begin{itemize}
\item
Let us assume that there is a weight rule
$r=\renh{h}\IF\limit{w_1}{\renh{\hid{A}}=W_{\renh{\hid{A}}},
                          \vis{A}=W_{\vis{A}}}
 \in\GLred{\eqt(P,Q)}{I}$
where $w_1=\max(0,w-\wsum{N}{\naf B=W_B})$ is associated with
$h\IF\limit{w}{A=W_A,\naf B=W_B}\in Q$ satisfying $h\in\hbh{Q}$.
By assuming $K\not\models r$, we obtain $K\not\models\renh{h}$ and
$w_1\leq\wsum{K}{\renh{\hid{A}}=W_{\renh{\hid{A}}},\vis{A}=W_{\vis{A}}}$.
It follows that
$w\leq
 \wsum{K}{\renh{\hid{A}}=W_{\renh{\hid{A}}},\vis{A}=W_{\vis{A}}}+
 \wsum{N}{\naf B=W_B}$
by the definition of $w_1$. On the other hand, the hidden part
$\eval{\hid{Q}}{\vis{M}}$ contains a weight rule
$h\IF\limit{w_2}{\hid{A}=W_{\hid{A}},\naf\hid{B}=W_{\hid{B}}}$
where
$w_2=
 \max(0,w-\wsum{\vis{M}}
               {\vis{A}=W_{\vis{A}},\naf\vis{B}=W_{\vis{B}}})$.
Thus the reduct $\GLred{(\eval{\hid{Q}}{\vis{M}})}{\hid{N}}$ contains a
rule $r'=h\IF\limit{w_3}{\hid{A}=W_{\hid{A}}}$ where the limit
$w_3=\max(0,w_2-\wsum{\hid{N}}{\naf\hid{B}=W_{\hid{B}}})$.
Using the definition of $w_2$, $N=\vis{M}\union\hid{N}$ and
$K$, we obtain $K\isect\hbv{Q}=\vis{M}$ and
from the preceding inequality concerning $w$,
$w_2\leq
 \wsum{K}{\renh{\hid{A}}=W_{\renh{\hid{A}}}}+
 \wsum{\hid{N}}{\naf\hid{B}=W_{\hid{B}}}$.
Similarly, the definition of $w_3$, yields us
$w_3\leq\wsum{K}{\renh{\hid{A}}=W_{\renh{\hid{A}}}}$.
But then the definition of $K$ implies that $r'$ is not satisfied by
$\lm{\GLred{(\eval{\hid{Q}}{\vis{M}})}{\hid{N}}}$, a contradiction.

\item
Suppose there is a rule $r=\renh{h}\IF\renh{\hid{A}},\vis{A}
\in\GLred{\eqt(P,Q)}{I}$ associated with a choice rule $\choice{H}\IF
A,\naf B\in Q$ such that $h\in\hid{H}$, $\hid{N}\models h$, and
$N\models\naf B$. Assuming $K\not\models r$ implies
$K\not\models\renh{h}$, $K\models\renh{\hid{A}}$, and
$K\models\vis{A}$. Since $K\isect\hbv{Q}=\vis{M}$
and $N=\vis{M}\union\hid{N}$ by definition, we know
that $\vis{M}\models\vis{A}\union\naf\vis{B}$.
Since
$\hid{H}\neq\emptyset$, it follows that
$\choice{\hid{H}}\IF\hid{A},\naf\hid{B}$ is
included in $\eval{\hid{Q}}{\vis{M}}$.
Moreover, the rule $r'=h\IF\hid{A}$ belongs to
$\GLred{(\eval{\hid{Q}}{\vis{M}})}{\hid{N}}$
as $\hid{N}\models\naf\hid{B}$ and $\hid{N}\models h$.
But then the definition of $K$ implies that
$r'$ is not satisfied by 
$\lm{\GLred{(\eval{\hid{Q}}{\vis{M}})}{\hid{N}}}$,
a contradiction.
\end{itemize}
The other rule types are covered by weight rules. It follows
by the structure of $\GLred{\eqt(P,Q)}{I}$ described in Lemma
\ref{lemma:reduct} that $K\models\GLred{\eqt(P,Q)}{I}$. In particular
the rules in Items
\ref{item:reduced-Q-begin}--\ref{item:reduced-rest-end}
are trivially satisfied by $K$ as their heads are.
It follows that $J\subseteq K$ and
$J\isect\renh{\hbh{Q}}\subseteq
 \renh{\lm{\GLred{(\eval{\hid{Q}}{\vis{M}})}{\hid{N}}}}$
as
$K\isect\renh{\hbh{Q}}=\renh{\lm{\GLred{(\eval{\hid{Q}}{\vis{M}})}{\hid{N}}}}$.
\vspace{1\baselineskip}

\noindent
\textbf{Claim \ref{item:least-model-of-reduct3}:}
If (i) and (ii), then $J\isect\ren{\hb{Q}}=\ren{\lm{\GLred{Q}{N}}}$.
\vspace{1\baselineskip}

Let us assume both (i) and (ii). It follows by Claims
\ref{item:least-model-of-reduct1} and \ref{item:least-model-of-reduct2}
that
$M=I\isect\hb{P}=\lm{\GLred{P}{M}}=J\isect\hb{P}$ and
$\renh{\hid{N}}=I\isect\renh{\hbh{Q}}=
 \renh{\lm{\GLred{(\eval{\hid{Q}}{\vis{M}})}{\hid{N}}}}=
 J\isect\renh{\hbh{Q}}$.

($\supseteq$)
Let us first establish $J\models\ren{(\GLred{Q}{N})}$.  It is clear by
Lemma \ref{lemma:reduct} that almost all rules of
$\ren{(\GLred{Q}{N})}$ are present in $\GLred{\eqt(P,Q)}{I}$.  The
only exception concerns a rule $r=\ren{h}\IF\ren{A}\union\set{h}$
(resp.\ $r=\ren{h}\IF\ren{A}\union\set{\renh{h}}$) included in
$\GLred{\eqt(P,Q)}{I}$ for a choice rule $\choice{H}\IF A,\naf B\in Q$
such that $h\in\vis{H}$ (resp.\ $h\in\hid{H}$) and $N\models\naf B$.
Suppose that $J\not\models r'$ for the corresponding rule
$r'=\ren{h}\IF\ren{A}$ included in $\ren{(\GLred{Q}{N})}$ which
presumes that $N\models h$. This implies $J\models h$ (resp.\
$J\models\renh{h}$) as $N=\vis{M}\union\hid{N}$ and $M=J\isect\hb{P}$
(resp. $\renh{\hid{N}}=J\isect\hbh{Q}$).  Thus $J\not\models r$, a
contradiction.
Hence $J\models\ren{(\GLred{Q}{N})}$ and
$J\isect\ren{\hb{Q}}\models\ren{(\GLred{Q}{N})}$.

($\subseteq$)
Let us then define an interpretation
$K=\lm{\GLred{P}{M}}\union
   \renh{\lm{\GLred{(\eval{\hid{Q}}{\vis{M}})}{\hid{N}}}}\union
   \ren{\lm{\GLred{Q}{N}}}\union
   \set{c,d,e}$.
It can be shown as in Claim \ref{item:least-model-of-reduct2} that
$K\models\GLred{P}{M}$ and the rules mentioned in Items
\ref{item:reduced-Q-h-begin}--\ref{item:reduced-Q-h-end}
of Lemma \ref{lemma:reduct} are satisfied by $K$.
As noted already, most of the rules of $\ren{(\GLred{Q}{N})}$ are
included in $\GLred{\eqt(P,Q)}{I}$ as such and thus satisfied by the
definition of $K$ as $\ren{\lm{\GLred{Q}{N}}}\models\ren{(\GLred{Q}{N})}$.
The only exceptions are made by rules $r$ of the forms defined above.
Suppose that $K\not\models r$ and define $r'=\ren{h}\IF\ren{A}$.  It
follows that $K\not\models r'$ and $h\in\lm{\GLred{P}{M}}$ (resp.\
$h\in\lm{\GLred{(\eval{\hid{Q}}{\vis{M}})}{\hid{N}}}$).  Then
$M=\lm{\GLred{P}{M}}$ (resp.\
$\hid{N}=\lm{\GLred{(\eval{\hid{Q}}{\vis{M}})}{\hid{N}}}$) implies
$N\models h$ so that $r'\in\ren{(\GLred{Q}{N})}$. Thus $N\models r'$
by the definition of $N$, a contradiction.
Finally, the rules in Items
\ref{item:reduced-rest-begin}--\ref{item:reduced-rest-end}
of Lemma \ref{lemma:reduct}
are satisfied by $K$ as $K\models\set{c,d,e}$.  Thus
$K\models\GLred{\eqt(P,Q)}{I}$. Since
$K\isect\ren{\hb{Q}}=\ren{\lm{\GLred{Q}{N}}}$, we obtain
$J\isect\ren{\hb{Q}}\subseteq\ren{\lm{\GLred{Q}{N}}}$.
\vspace{1\baselineskip}

\noindent
\textbf{Claim \ref{item:least-model-of-reduct4}:}
If (i), (ii), (iii), and $A=J\isect\set{c,d,e}$, then
(a) $d\in A$ $\iff$ $N\neq L$,
(b) $c\in A$ $\iff$ $d\not\in I$ and $L\not\models\compst{Q}$, and
(c) $e\in A$ $\iff$ $c\in A$ or $d\in A$.
\vspace{1\baselineskip}

Assume (i), (ii), and (iii). Using Claims
\ref{item:least-model-of-reduct1}--\ref{item:least-model-of-reduct3},
we obtain
$M=I\isect\hb{P}=\lm{\GLred{P}{M}}=J\isect\hb{P}$,
$\renh{\hid{N}}=I\isect\renh{\hbh{Q}}=
 \renh{\lm{\GLred{(\eval{\hid{Q}}{\vis{M}})}{\hid{N}}}}=
 J\isect\renh{\hbh{Q}}$, and
$\ren{L}=I\isect\ren{\hb{Q}}=
 \ren{\lm{\GLred{Q}{N}}}=
 J\isect\ren{\hb{Q}}$.

\begin{itemize}
\item[(a)]
The structure of $\GLred{\eqt(P,Q)}{I}$ made explicit in Lemma
\ref{lemma:reduct} and the properties of $\lm{\GLred{\eqt(P,Q)}{I}}$
imply that $d\in A$
$\iff$
there is an atom $a\in\hb{Q}$ such that $L\not\models a$ and $N\models
a$; or $N\not\models a$ and $L\models a$.  But this is equivalent to
$N\neq L$.

\item[(b)]
The same premises imply that $c\in A$
$\iff$
$c\in J$
$\iff$
$I\not\models d$; and
there is $a\in\compst{Q}$ such that $L\not\models a$ or
or there $\naf b\in\compst{Q}$ such that $L\models b$.
Or equivalently, $d\not\in I $ and $L\not\models\compst{Q}$.

\item[(c)]
Finally, we have $e\in A$
$\iff$
$J\models e$
$\iff$
$J\models c$ or $J\models d$
$\iff$ $c\in A$ or $d\in A$.
\end{itemize}
\end{proof}


\begin{proof}[Proof of Theorem \ref{theorem:correctness}]
($\implies$)
Suppose that $\eqt(P,Q)$ has a stable model $K$, i.e.\
$K=\lm{\GLred{\eqt(P,Q)}{K}}$ and $K\models\compst{\eqt(P,Q)}$.
Let us then extract three interpretations from $K$:
$M=K\isect\hb{P}$, $N=\vis{M}\union\hid{N}$ where
$\hid{N}=\sel{a\in\hbh{Q}}{\renh{a}\in K}$, and
$L=\sel{a\in\hb{Q}}{\ren{a}\in K}$.
It follows that $M=K\isect\hb{P}=\lm{\GLred{P}{M}}$ by Claim
\ref{item:least-model-of-reduct1} in Proposition
\ref{prop:least-model-of-reduct}. Besides, we have
$M\models\compst{P}$ as $K\models\compst{\eqt(P,Q)}$ and
$\compst{P}\subseteq\compst{\eqt(P,Q)}$. Thus $M\in\sm{P}$.

We may now apply Claim \ref{item:least-model-of-reduct2} in Proposition
\ref{prop:least-model-of-reduct} since condition (i) is satisfied. Thus
$\renh{\hid{N}}=K\isect\renh{\hbh{Q}}=
 \renh{\lm{\GLred{(\eval{\hid{Q}}{\vis{M}})}{\hid{N}}}}$
which makes condition (ii) true in Proposition \ref{prop:least-model-of-reduct}
so that $\hid{N}\in\sm{\eval{\hid{Q}}{\vis{M}}}$ is the case.

This enables the use of Claim \ref{item:least-model-of-reduct3} in
Proposition \ref{prop:least-model-of-reduct} to obtain
$\ren{L}=K\isect\ren{\hb{Q}}=\ren{\lm{\GLred{Q}{N}}}$. Thus
$L=\lm{\GLred{Q}{N}}$ and condition (iii) in Proposition
\ref{prop:least-model-of-reduct} is satisfied.

On the other hand, $e\in A$ holds for $A=K\isect\set{c,d,e}$ as
$K\models\compst{\eqt(P,Q)}$ and $e\in\compst{\eqt(P,Q)}$ by
Definition \ref{def:translation}. It follows by (c) and (b) in Claim
\ref{item:least-model-of-reduct4} of Proposition
\ref{prop:least-model-of-reduct} that $c\in A$ or $d\in A$, i.e.\
$d\not\in A$ and $L\not\models\compst{Q}$; or $d\in A$. Using (a) we
obtain $N=L$ and $L\not\models\compst{Q}$; or $N\neq L$. By
substituting $\lm{\GLred{Q}{N}}$ for $L$ and $N$ for $L$, we have
$N=\lm{\GLred{Q}{N}}$ and $N\not\models\compst{Q}$; or
$N\neq\lm{\GLred{Q}{N}}$. Since $Q$ has enough visible atoms, we know
that $\hid{N}$ is unique with respect to $Q$ and $\vis{M}$, and
there is no $N\in\sm{Q}$ such that $\vis{N}=\vis{M}$.

($\impliedby$)
Suppose that $P$ has a stable model $M=\lm{\GLred{P}{M}}$ and there
is no $N\in\sm{Q}$ such that $\vis{N}=\vis{M}$. Since $Q$ has enough
visible atoms any such candidate $N$ must be based on the unique
stable model $\hid{N}=\lm{\GLred{(\eval{\hid{Q}}{\vis{M}})}{\hid{N}}}$.
So let us define $N=\vis{M}\union\hid{N}$.
The instability of $N$ implies either $N\neq\lm{\GLred{Q}{N}}$; or
$N=\lm{\GLred{Q}{N}}$ and $N\not\models\compst{Q}$. In either case,
let $L=\lm{\GLred{Q}{N}}$. Moreover, let $A\subseteq\set{c,d,e}$ be a set
of atoms so that $d\in A$ $\iff$ $N\neq\lm{\GLred{Q}{N}}$, $c\in A$
$\iff$ $N=\lm{\GLred{Q}{N}}$ and $N\not\models\compst{Q}$, and $e\in
A$ unconditionally.

Let us then define an interpretation
$K=M\union\renh{\hid{N}}\union\ren{L}\union A$.
It is easy to see that $K\models\compst{\eqt(P,Q)}$ as
$M\models\compst{P}$ and $K\models e$ by definition.  It remains to
establish that $K=\lm{\GLred{\eqt(P,Q)}{K}}$.
First, the definition of $K$ implies that $K\isect\hb{P}=M$.  It
follows by Claim \ref{item:least-model-of-reduct1} in Proposition
\ref{prop:least-model-of-reduct} that
$\lm{\GLred{\eqt(P,Q)}{K}}\isect\hb{P}=\lm{\GLred{P}{M}}=M$.
Second, we have $K\isect\renh{\hbh{Q}}=\renh{\hid{N}}$ by definition.
Using Claim \ref{item:least-model-of-reduct2} in Proposition
\ref{prop:least-model-of-reduct} we obtain
$\lm{\GLred{\eqt(P,Q)}{K}}\isect\renh{\hbh{Q}}=
 \renh{\lm{\GLred{(\eval{\hid{Q}}{\vis{M}})}{\hid{N}}}}=\renh{\hid{N}}$.
Third, we defined $K$ so that $K\isect\ren{\hb{Q}}=\ren{L}$. It follows by
Proposition \ref{prop:least-model-of-reduct} (Claim
\ref{item:least-model-of-reduct3}) that
$\lm{\GLred{\eqt(P,Q)}{K}}\isect\ren{\hb{Q}}=\ren{\lm{\GLred{Q}{N}}}=\ren{L}$.
Finally, we recall that $K\isect\set{c,d,e}=A$. It follows by Claim
\ref{item:least-model-of-reduct4} in Proposition
\ref{prop:least-model-of-reduct} that
(a)
$d\in\lm{\GLred{\eqt(P,Q)}{K}}$ $\iff$ $N\neq L$ $\iff$
$N\neq\lm{\GLred{Q}{N}}$ $\iff$ $d\in A$ by the definition of $A$ above;
(b)
$c\in\lm{\GLred{\eqt(P,Q)}{K}}$ $\iff$ $d\not\in K$ and
$L\not\models\compst{Q}$ $\iff$ $d\not\in A$ and
$L\not\models\compst{Q}$ $\iff$ $N=L$ and
$N\not\models\compst{Q}$ $\iff$ $N=\lm{\GLred{Q}{N}}$
and $N\not\models\compst{Q}$ $\iff$ $c\in A$; and
(c)
$e\in\lm{\GLred{\eqt(P,Q)}{K}}$ holds as the instability of $N$
implies either $d\in A$ or $c\in A$.
Thus $\lm{\GLred{\eqt(P,Q)}{K}}\isect\set{c,d,e}=A$.
To summarize, we have established
$\lm{\GLred{\eqt(P,Q)}{K}}=M\union\renh{\hid{N}}\union\ren{L}\union
A=K$.  Thus $K\in\sm{\eqt(P,Q)}$.
\end{proof}


\begin{proof}[Proof of Proposition \ref{prop:same-semantics}]
Let $M\subseteq\hb{P}$ be any interpretation for $P$ and $P_w$.
We rewrite (\ref{eq:choice-as-wcr}) using shorthands as
$\limit{0}{H=\mathbf{1}}\IF\limit{|A|+|B|}{A=\mathbf{1},\naf B=\mathbf{1}}$
where $\mathbf{1}$s are sets of weights of appropriate sizes
consisting of only $1$s. As regards the respective choice rule
$\choice{H}\IF A,\naf B$ and any $h\in H$, Definition \ref{def:reduct}
implies that $h\IF A$ belongs to $\GLred{P}{M}$ $\iff$ $M\models h$
and $M\models\naf B$.  On the other hand,
Definition \ref{def:wc-program-reduct} implies
$h\IF\limit{(|A|+|B|-\wsum{M}{\naf B=\mathbf{1}})}{A=\mathbf{1}}
 \in\GLred{P_w}{M}$
$\iff$ $M\models h$.
Quite similarly, we use $\limit{1}{h=1}\IF\limit{w}{A=W_A,\naf B=W_B}$
as an abbreviation for (\ref{eq:weight-as-wcr}). Then the reduced
rule $h\IF\limit{w'}{A=W_A}$ where $w'=\max(0,w-\wsum{M}{\naf B=W_B})$
belongs to $\GLred{P}{M}$ unconditionally and to $\GLred{P_w}{M}$
$\iff$ $M\models h$.

$(\implies)$
Suppose that $M=\lm{\GLred{P}{M}}$. It follows immediately that
$M\models\GLred{P}{M}$ and $M\models P$. Since choice rules and their
translations (\ref{eq:choice-as-wcr}) do not interfere with the
satisfaction of rules, we conclude $M\models P_w$ by the close
relationship of (\ref{eq:weight-rule}) and (\ref{eq:weight-as-wcr}).
Moreover, it is easy to see that $\lm{\GLred{P_w}{M}}\subseteq M$ as
the analysis above shows that the head atom $h$ of every rule included in
$\GLred{P_w}{M}$ is necessarily true in $M$, i.e., $h\in M$.

It remains to prove by induction that each interpretation in a
sequence defined by $M_0=\emptyset$ and
$M_i=\nec{\GLred{P}{M}}(M_{i-1})$ for $i>0$ is contained in
$\lm{\GLred{P_w}{M}}$. Note that $M_i\subseteq M$ for each $i\geq 0$
and $M=\lfp{\nec{\GLred{P}{M}}}=M_i$ for some finite $i$ due to
compactness of $\nec{\GLred{P}{M}}$.
Let us the consider any $h\in M_i$. Note that $h\in M$ holds, i.e.,
$M\models h$. The definition of $M_i$ implies that (i) there is a rule
$h\IF A\in\GLred{P}{M}$ such that $M\models\naf B$ and $A\subseteq
M_{i-1}$; {\em or} (ii) there is a rule
$h\IF\limit{w'}{A=W_A}\in\GLred{P}{M}$ with
$w'\leq\wsum{M_{i-1}}{A=W_A}$.
If (i) holds, the rule $h\IF\limit{|A|}{A=\mathbf{1}}$ belongs to
$\GLred{P_w}{M}$ as $M\models h$. Moreover, $A\subseteq
M_{i-1}\subseteq\lm{\GLred{P_w}{M}}$ by induction hypothesis. In
case of (ii), $M\models h$ implies that the reduced rule is also in
$\GLred{P_w}{M}$. Since $M_{i-1}\subseteq\lm{\GLred{P_w}{M}}$, we
obtain $w'\leq\wsum{\lm{\GLred{P_w}{M}}}{A=W_A}$.
Thus $h\in\lm{\GLred{P_w}{M}}$ results in both cases so that
$M_i\subseteq\lm{\GLred{P_w}{M}}$ for each $M_i$ and
$M$ in particular so that $M=\lm{\GLred{P_w}{M}}$.

($\impliedby$)
Let us then assume that $M\models P_w$ and $M=\lm{\GLred{P_w}{M}}$ as
well as $M\not\models\GLred{P}{M}$. The last cannot be caused by a
choice rule because $h\IF A$ is included in $\GLred{P}{M}$ only if
$M\models h$. If a weight rule is the reason, then
$h\IF\limit{w'}{A=W_A}$ with $w'=\max(0,w-\wsum{M}{\naf B=W_B})$
belongs to $\GLred{P}{M}$, $w'\leq\wsum{M}{A=W_A}$, and $M\not\models
h$. By adding $\wsum{M}{\naf B=W_B}$ on both sides of the inequality,
we obtain $w\leq\wsum{M}{A=W_A,\naf B=W_B}$. Thus a rule
$\limit{1}{h=1}\IF\limit{w}{A=W_A,\naf B=W_B}$ of $P_w$ is not
satisfied by $M$, a contradiction. Hence
$M\models\GLred{P}{M}$.

Now $M\models\GLred{P}{M}$ implies $\lm{\GLred{P}{M}}\subseteq M$ and
we need induction to establish inclusion in the other direction. This
time we use a sequence defined by $M_0=\emptyset$ and
$M_i=\nec{\GLred{P_w}{M}}(M_{i-1})$ for $i>0$. Then consider any $h\in
M_i$. Since $M$ is the limit of the sequence, we obtain $h\in M$ and
$M\models h$. Moreover, the definition of $M_i$ implies that (iii) there
is a rule $h\IF\limit{w''}{A=\mathbf{1}}\in\GLred{P_w}{M}$ where
$w''=|A|+|B|-\wsum{M}{\naf
B=\mathbf{1}}\leq\wsum{M_{i-1}}{A=\mathbf{1}}$; or (iv) there is a
rule $h\IF\limit{w'}{A=W_A}\in\GLred{P_w}{M}$ such that
$w'\leq\wsum{M_{i-1}}{A=W_A}$.
In case of (iii), we infer $\wsum{M}{\naf B=\mathbf{1}}=|B|$ and
$\wsum{M_{i-1}}{A=\mathbf{1}}=|A|$ as necessities so that
$M\models\naf B$ and $A\subseteq M_{i-1}$ follow. Thus $h\IF
A\in\GLred{P}{M}$ as $M\models h$ and $\lm{\GLred{P}{M}}\models A$
follows by the induction hypothesis $M_{i-1}\subseteq\lm{\GLred{P}{M}}$.
If (iv) holds, the reduced rule is also a member of
$\GLred{P}{M}$ by definition. Using the induction hypothesis again,
we obtain $w'\leq\wsum{\lm{\GLred{P}{M}}}{A=W_A}$.
To conclude the preceding case analysis, we have
$h\in\lm{\GLred{P}{M}}$ for any $h\in M_i$ and thus
$M_i\subseteq\lm{\GLred{P}{M}}$. Since $M=M_i$ for some $i$, we obtain
$M\subseteq\lm{\GLred{P}{M}}$.
\end{proof}


\begin{proof}[Proof of Theorem \ref{theorem:tr-sns-faithful}]
Consider any weight constraint program $P$. Now $P\lpeq{v}\tr{SNS}{P}$
holds by the definition of $\lpeq{v}$ if and only if
$\hbv{P}=\hbv{\trop{}(P)}$ and there is a bijection
$\ext\func{\sm{P}}{\sm{\tr{SNS}{P}}}$ such that for all $M\in\sm{P}$
it holds that $M\isect\hbv{P}=\ext(M) \isect\hbv{\tr{SNS}{P}}$. Since
$\hbv{P}=\hbv{\tr{SNS}{P}}$ holds by Definition
\ref{def:tr-wcp-for-smodels}, it remains to to establish such a
bijection $\ext$ from $\sm{P}$ to $\sm{\tr{SNS}{P}}$.

Given an interpretation $M\subseteq\hb{P}$, we define $\ext(M)=M\union
\sus{P}{M}$ where $\sus{P}{M}$ satisfies for each weight constraint
$C=l\leq\{A=W_A,\naf B=W_B\}\leq u$ appearing in $P$ that

\begin{enumerate}
\item
$\SAT{C}\in\sus{P}{M}$ $\iff$
$l\leq\wsum{M}{A=W_{A},\naf B=W_{B}}$, and
\item
$\UNSAT{C}\in\sus{P}{M}$ $\iff$ $u+1\leq\wsum{M}{A=W_{A},\naf B=W_{B}}$.
\end{enumerate}

Now, if $M\in\sm{P}$, then $N=\ext(M)\in\sm{\tr{SNS}{P}}$ follows by
the results of \citeN{SNS02:aij}. Thus $\ext$ is indeed a function from
$\sm{P}$ to $\sm{\tr{SNS}{P}}$ and it remains to establish that
$\ext$ is a bijection. It is clearly injective as $M_1\not=M_2$
implies $\ext(M_1)\not=\ext(M_2)$ by the definition of $\ext$.

To prove that $\ext$ is also a surjection, let us consider any
$N\in\sm{\tr{SNS}{P}}$ and the respective projection
$M=N\isect\hb{P}$.
Since $N\in\sm{\tr{SNS}{P}}$, it holds that $N\models\tr{SNS}{P}$ and
moreover $M\in\sm{P}$ holds \cite{SNS02:aij}. Thus we need to show
$N=N'$ for
$N'=\ext(M)=M\union\sus{P}{M}$.
Since $\ext\func{\sm{P}}{\sm{\tr{SNS}{P}}}$ we know that
$N'\in \sm{\tr{SNS}{P}}$.

Let us show that $\sus{P}{M}\subseteq N$.  Assuming the opposite there
is an atom $a\in\sus{P}{M}$ such that $a\not\in N$. By the definition
of $\sus{P}{M}$ either (i) $a=\SAT{C}$ or (ii) $a=\UNSAT{C}$ for some
$C=l\leq\{A=W_{A},\naf B=W_{B}\}\leq u$ appearing in $P$. This leads
to a case analysis as follows.

\begin{enumerate}
\item[(i)]
If $a=\SAT{C}\in\sus{P}{M}$, then there is a rule (\ref{eq:lower-bound})
in $\tr{SNS}{P}$ such that
$l\leq
 \wsum{M}{A=W_{A},\naf B=W_{B}}=
 \wsum{N}{A=W_{A},\naf B=W_{B}}$
where last equality holds by the definition of $M$ as
$A\subseteq\hb{P}$ and $B\subseteq\hb{P}$. Since $\SAT{C}\not\in
N$, it follows that (\ref{eq:lower-bound}) is not satisfied by
$N$. But this contradicts $N\models\tr{SNS}{P}$.

\item[(ii)]
Quite similarly, if $a=\UNSAT{i}\in\sus{P}{M}$, then there is a rule
(\ref{eq:upper-bound}) such that
$u+1\leq
 \wsum{M}{A=W_{A},\naf B=W_{B}}=
 \wsum{N}{A=W_{A},\naf B=W_{B}}$.
Then (\ref{eq:upper-bound}) is not satisfied by $N$ as
$\UNSAT{i}\not\in N$. A contradiction with $N\models\tr{SNS}{P}$.
\end{enumerate}

Hence $\sus{P}{M}\subseteq N$ is necessarily the case. Since $M\subseteq
N$ by definition, we have $N'\subseteq N$. How about the converse
inclusion $N\subseteq N'=M\union\sus{P}{M}$? It is clear that
$N\isect\hb{P}=M\subseteq N'$. Then a potential difference
$N'\setminus N$ (if any) must be caused by new atoms involved in
$\tr{SNS}{P}$. There are three kinds of such atoms.
\begin{enumerate}
\item
Suppose that $\SAT{C}\in N$ for some $C=l\leq\{A=W_A,\naf B=W_B\}\leq
u$ appearing in $P$.  Since $N$ is a stable model of $\tr{SNS}{P}$ and
there is only one rule (\ref{eq:lower-bound}) in $\tr{SNS}{P}$ having
$\SAT{C}$ as its head, the body of that rule must be satisfied in $N$,
too, i.e.,
$l\leq\wsum{N}{A=W_{A},\naf B=W_{B}}$.
Since $M=N\isect\hb{P}$, $A\subseteq\hb{P}$, and $B\subseteq\hb{P}$,
the same holds for $M$. Thus $\SAT{C}\in \sus{P}{M}$.

\item
Using the same line of reasoning and the rule (\ref{eq:upper-bound})
included in $\tr{SNS}{P}$, we know that $\UNSAT{C}\in N$ implies
$\UNSAT{C}\in\sus{P}{M}$.

\item
Now $f\not\in N$ must hold as $N$ is a stable model of $\tr{SNS}{P}$
which includes (\ref{eq:check-head3}).
\end{enumerate}

To conclude, we have established $N\subseteq N'$ which indicates that
there is $M\in\sm{P}$ such that $N=\ext(M)$. Therefore $\ext$ is
bijective and $\trop{SNS}$ faithful.
\end{proof}


\end{document}